\documentclass[lettersize,journal]{IEEEtran}

\usepackage{amsmath}
\usepackage{amssymb}
\usepackage{mathtools} 
\usepackage{tabularx} 
\usepackage{threeparttable}
\usepackage{booktabs}
\usepackage{graphicx}
\usepackage[vlined, ruled, linesnumbered, noend]{algorithm2e} 
\usepackage[normalem]{ulem}
\usepackage{subcaption}
\usepackage{eqlist}
\usepackage{amsfonts} 
\usepackage{bbm} 
\usepackage{enumitem}
\setlist[itemize]{noitemsep, topsep=0pt}
\usepackage{lipsum}
\usepackage{soul}
\usepackage{amsthm}

\usepackage[usenames,dvipsnames]{xcolor}
\usepackage{algpseudocode}
\usepackage{tcolorbox}
\usepackage{comment}
\usepackage{url}

\usepackage{multirow}

\usepackage{tikz}
\usetikzlibrary{fit,shapes.misc}

\usepackage[hang,flushmargin]{footmisc}
\usepackage{lipsum}
\makeatletter
\newcommand{\algorithmfootnote}[2][\footnotesize]{%
  \let\old@algocf@finish\@algocf@finish
  \def\@algocf@finish{\old@algocf@finish
    \leavevmode\rlap{\begin{minipage}{\linewidth}
    #1#2
    \end{minipage}}%
  }%
}

\usepackage{CJKutf8}
\definecolor{bleudefrance}{rgb}{0.19, 0.55, 0.91}
\definecolor{awesome}{rgb}{1.0, 0.13, 0.32}

\definecolor{darkgreen}{rgb}{0.0, 0.65, 0.0}

\newtheorem*{problem}{Problem Definition}

\begin{document}
\title{Robotic Test Tube Rearrangement Using Combined Reinforcement Learning and Motion Planning}
\author{Hao Chen$^{1}$, Weiwei Wan$^{1*}$, Masaki Matsushita$^2$, Takeyuki Kotaka$^2$ and Kensuke Harada$^{1,3}$
\thanks{$^{1}$Graduate School of Engineering Science, Osaka University, Japan. $^{2}$H.U. Group Research Institute G.K., Japan. $^{3}$National Inst. of AIST, Japan. Contact: Weiwei Wan, {\tt\small wan@hlab.sys.es.osaka-u.ac.jp}}
}

\markboth{Preprint Version}
{Chen \MakeLowercase{\textit{et al.}}: Robotic Test Tube Rearrangement Using Combined Reinforcement Learning and Motion Planning} 
\maketitle

\begin{abstract}
A combined task-level reinforcement learning and motion planning framework is proposed in this paper to address a multi-class in-rack test tube rearrangement problem. At the task level, the framework uses reinforcement learning to infer a sequence of swap actions while ignoring robotic motion details. At the motion level, the framework accepts the swapping action sequences inferred by task-level agents and plans the detailed robotic pick-and-place motion. The task and motion-level planning form a closed loop with the help of a condition set maintained for each rack slot, which allows the framework to perform replanning and effectively find solutions in the presence of low-level failures. Particularly for reinforcement learning, the framework leverages a distributed deep Q-learning structure with the Dueling Double Deep Q Network (D3QN) to acquire near-optimal policies and uses an A${}^\star$-based post-processing technique to amplify the collected training data. The D3QN and distributed learning help increase training efficiency. The post-processing helps complete unfinished action sequences and remove redundancy, thus making the training data more effective. We carry out both simulations and real-world studies to understand the performance of the proposed framework. The results verify the performance of the RL and post-processing and show that the closed-loop combination improves robustness. The framework is ready to incorporate various sensory feedback. The real-world studies also demonstrated the incorporation.
\end{abstract}


\section{Introduction}
\label{sec:introduction}

\IEEEPARstart{R}{earrangement} planning is an active and challenging research area in robotics that typically involves finding a sequence of motion to rearrange objects from an initial configuration to a goal configuration. In this paper, we focus on a multi-class in-rack test tube rearrangement, which requests rearranging various test tubes randomly placed within a rack into a specified pattern.
The problem is challenging due to constraints such as intricate mutual blockage among the tubes, difficulty in carrying out replanning in the presence of motion planning and execution failures, etc. Additionally, the complexity of the rearrangement procedure can escalate, leading to an increased number of required motions, extended time, higher energy consumption, and susceptibility to disruptions that compromise overall success. Given these challenges, we propose a combined task-level reinforcement learning and motion planning framework to address the multi-class in-rack test tube rearrangement problem. At the task level, the framework leverages a Dueling Double Deep Q Network (D3QN) \cite{van2016deep}\cite{wang2016dueling} and distributed Q-learning \cite{mnih2015human} to acquire near-optimal policies for inferring action sequences while ignoring robotic motion details. At the motion level, the framework accepts the inferred task-level sequences and plans the robotic pick-and-place motion in the robot configuration space for implementing the sequence. The task-level RL inference and the motion-level planning are combined in a closed loop by maintaining a condition set for each rack slot.


Previously, many researchers utilized combined task and motion planners to solve rearrangement planning problems. The task-level planner usually employed heuristic search (e.g., A${}^\star$ \cite{lee2021tree} or Monte Carlo Tree Search (MCTS) \cite{labbe2020monte}) to plan feasible action sequences. The heuristic methods are ineffective as most rearrangement problems are NP-hard \cite{labbe2020monte}. As the type and number of objects increase, heuristics become limited in scope and require significant computation \cite{lee2021tree}\cite{wang2022efficient}. Recent advances in reinforcement learning (RL) provide an alternative approach for task planning, which enables autonomous exploration of rearrangement sequences through iterative trial-and-error. By learning and continuously improving a policy from experience, RL agents can overcome the limited planning scope of heuristics. The trained models also have higher inference efficiency compared to heuristic methods. However, the sample complexity of RL may grow quadratically with the learning horizon \cite{dann2015sample}. This makes its application to rearrangement problems challenging, as such problems require long-term iteration and may cause RL agents to demand substantially large amounts of interactions and computational resources for convergence \cite{cheng2021heuristic}. Taking into account both the strengths and weaknesses, we use RL as the task-level planner in our framework, and at the same time, we develop an A${}^\star$-based post-processing technique to amplify the collected training data. The post-processing enhances the effectiveness of the generated training data by completing unfinished action sequences and eliminating redundancy, thereby improving training performance.

The motion planners usually ran subsequently after the task planner in the previous studies. They were designed to accept task sequences and plan the detailed robotic pick-and-place motion in the robot configuration space. When motion planning fails, previous studies employed a simple closed-loop strategy to replan between task and motion levels. They invalidated the primary action and selected a secondary feasible action for exploration \cite{schiavi2023learning}. While effective in certain situations, the method faces challenges in environments with intricate dependencies and large action space. The reason is that actions are closely tied to states. As the state changes, similar actions may arise and cause replanning, leading to significant time consumption and even infinity. To address the challenges, we propose maintaining a condition set for each rack slot and selecting actions by considering constraints formed by the conditions tied to respective slots. The condition sets allow us to perform replanning and effectively find solutions in the presence of motion planning failures.

In summary, our work has two main contributions:
\begin{enumerate} 
    \item Our task planner leverages D3QN and distributed Q-learning to infer action sequences. To improve training performance and assure convergence, we develop an A${}^\star$-based post-processing technique to amplify the collected training data. The post-processing helps complete unfinished action sequences and remove redundancy, thus making the training data more effective.
    \item We close the loop of the task and motion level planner by maintaining a condition set for each rack slot. The sets allow us to select actions considering the constraints formed by the conditions tied to respective slots, thus efficiently performing replanning and effectively finding solutions in the presence of motion failures.
\end{enumerate}

In addition, our framework is ready to incorporate various sensory feedback like gripper jaw width, vision, and torque/force (current) for robust execution. The failures detected by sensors can be solved similarly within the closed loop. We carry out both simulations and real-world studies to understand the performance of the proposed framework. The results demonstrate the efficacy of the RL and post-processing, verify the robustness and performance improvements enabled by our closed-loop combination, and show the pragmatic flavor of our framework in the real world.

\section{Related Work}
\label{sec:related_work}

\subsection{Rearrangement Planning}
Rearrangement planning is a widely studied topic in robotics encompassing various problems such as tabletop/room rearrangement \cite{labbe2020monte}\cite{weihs2021visual}, tabletop sorting \cite{huang2019large}, object retrieval within cluttered environments \cite{wang2022efficient}, as well as navigation or manipulation around movable obstacles \cite{stilman2005navigation}\cite{saxena2021manipulation}, etc. The general aim of these problems is to rearrange objects from an initial configuration to a desired goal configuration. Most of them involve both task-level and robot motion-level planning to find a feasible solution and have NP-hard complexity \cite{gao2023minimizing}. More recent studies on rearrangement planning shifted from finding feasible solutions to minimizing manipulations, which makes determining optimal rearrangement plans even more computationally intractable. From the task-level perspective, previous rearrangement planning can be categorized into conventional AI-based methods and learning-based methods. 

\subsubsection{Conventional AI-Based Methods}
Early rearrangement planning studies used sampling-based, graph-based search algorithms at the task level. For instance, King et al. \cite{king2016rearrangement} and Haustein et al. \cite{haustein2015kinodynamic} used Kinodynamic RRT to find a feasible non-prehensile motion for rearranging objects. Han et al. \cite{han2018complexity} formulated tabletop object rearrangement with overhand grasps as a Traveling Salesman Problem (TSP) and solved it using graph-based search while considering the travel distances of robots. Later work from the same group used a dependency graph to formulate rearrangement problems with buffer places \cite{gao2021on}. Inspired by multi-robot motion planning, Krontiris et al. \cite{krontiris2014rearranging} used pebble graphs to represent unlabeled rearrangement problems. Collision-free object poses were encoded as nodes of the graphs. Labb{\'e} et al. \cite{labbe2020monte} and Song et al. \cite{song2020multi} respectively utilized Monte Carlo Tree Search (MCTS) for arranging loosely placed tabletop objects and separating cluttered objects from various classes using push, respectively. 

Most of the early studies use a hierarchical structure, which first finds feasible solutions at the task level and then passes to a lower level for detailed robotic motion generation. Despite their effectiveness, conventional AI-based methods face challenges in efficiency and scalability. They have difficulty in solving complicated problems, e.g., problems with multiple objects, multiple goals, etc. Also, the methods heavily depend on the quality of the heuristic functions. Poorly designed heuristics can result in sub-optimal solutions or extended computation times. For these reasons, recent research has shifted to learning-based methods.

\subsubsection{Learning-Based Methods}
Both supervised and RL methods have gained popularity due to their distinct advantages. Supervised learning methods are popular for scenarios where action sequences are less critical or object interdependence is minor. For instance, Liu et al. \cite{liu2022structformer}, Paxton et al. \cite{paxton2022predicting}, and Qureshi et al. \cite{qureshi2021nerp} respectively demonstrated the effectiveness of supervised learning in moving a few unseen objects. Huang et al. \cite{huang2023planning} and Kulshrestha et al. \cite{kulshrestha2023structural} respectively used Graph Neural Networks (GNN) to analyze object relations and generate manipulation plans. In more complicated scenarios, particularly those requiring end-to-end visual-motor policies or involving multiple objects, RL-based methods are more advantageous. For example, Yuan et al. \cite{yuan2019end} and Tang et al. \cite{tang2022reinforcement} respectively used Deep Q-Networks (DQN) and Proximal Policy Optimization (PPO) to generate end-to-end visual-push motion for tabletop object rearrangement. Zeng et al. \cite{zeng2018learning} extended the two studies by incorporating grasping actions, and trained end-to-end policies capable of choosing between pushing and grasping actions in scenarios involving over 30 objects. Deng et al. \cite{deng2022deep} used GNN and DQN to solve deformable object rearrangement. Wang et al. \cite{wang2020scene} and Bai et al. \cite{bai2022hierarchical} respectively used DQN and PPO to learn a high-level policy to guide MCTS to find the most efficient action sequence for object arrangements. Cheong et al. \cite{cheong2021obstacle} used DQN to learn a policy in a lattice environment to relocate obstacles and thus retrieve target objects.

This study focuses on a multi-class in-rack test tube rearrangement task. Our previous research \cite{wan2022arranging} highlighted the limitations of heuristic-based search in achieving efficient responsiveness for such tasks. In this paper, we propose a combined task-level RL and motion planning framework to address these limitations. Our framework used D3QN and distributed learning to ensure efficient task-level learning and combined the trained policy with motion planning in a closed-loop manner to improve robustness.

\subsection{Heuristics and Data Relabeling in RL}
\subsubsection{Heuristic Acceleration in RL}
Many previous studies have employed heuristics as prior knowledge in RL to accelerate training. They can be divided into three categories based on how heuristics are used.

\paragraph{Pre-training on heuristically generated datasets} This category of methods pre-trains RL agents using datasets generated by heuristic search \cite{larsson2018evaluation}. The data sets help accelerate policy learning and the RL network could exhibit competitive performance when the learned policy is combined with value approximation updates \cite{da2021learning}.

\paragraph{Reward shaping using heuristics} This category of methods integrates heuristics into the RL reward function to expedite policy development. For example, Xie et al. \cite{xie2020heuristic} used a heuristic reward function throughout the RL process to tackle sparse rewards. The developed methods were applied to three-dimensional path planning for drones. Nonetheless, shaping the rewards throughout training narrows down exploration towards specific actions/states, and thus results in strong human bias. To reduce bias, Li and Xiang \cite{li2023vanishing} proposed using heuristic guidance primarily in the early training stages and progressively decreasing its influence as training advances. Cheng et al. \cite{cheng2021heuristic} proposed a more nuanced approach that used horizon-based regularization to determine whether long-term value information should come from collected experiences or the heuristics.

\paragraph{Heuristically guided exploration} Two paradigms are often used for heuristically guided exploration. The first one mathematically combines an estimation function and a heuristic function \cite{bianchi2008accelerating,morozs2015heuristically,esteves2021heuristically,ji2023heuristically} for action selection. The second one replaces random actions in the $\varepsilon$-greedy method with actions suggested by heuristics \cite{anh2007heuristic}\cite{luo2023guiding}.

The efforts and successes of the above publications confirmed that heuristic data is advantageous to RL. However, determining when and how to include the heuristic data for training and, at the same time, maintaining high scalability remains challenging. The heuristically generated data is susceptible to biases, and the effectiveness of exploration is significantly influenced by the heuristic functions. Also, the heuristically generated data often only covers a narrow part of the state space. The data would be inefficient for problems that lack high-quality demonstrations \cite{hester2018deep} or explicit measurements. 

Our way of incorporating heuristics in this work does not belong to any of the above-mentioned categories. We employ heuristics in a post-processing procedure to amplify the data, which is essentially an extension of the data relabeling concept discussed below.
 
\subsubsection{Data Relabeling in RL}
Data relabeling, particularly in the context of off-policy learning, has emerged as a powerful technique to enhance sample efficiency in RL. A noticeable example is Hindsight Experience Replay (HER) for goal-conditioned RL problems \cite{andrychowicz2017hindsight}, which relabeled past experiences with new goals, thereby transforming unsuccessful episodes into valuable learning experiences. Subsequent improvements to HER include methods like the adaptive selection failed experiences for relabeling \cite{fang2019curriculum}, multi-step hindsight experience replay \cite{yang2023multi}, etc. In parallel with the HER method and its extensions, Lee et al. \cite{lee2021pebble} proposed a relabeling framework that continuously adapted a reward function based on human preferences and updated rewards in all past experiences. While conventional data relabeling focused primarily on goals or rewards, our study extends this concept to states and actions, thereby extracting additional value from past experiences to further enhance data effectiveness. 

We call our extension the post-processing technique. It is based on the A$^\star$ algorithm presented in our previous work \cite{wan2022arranging}. For one thing, we employed the A${}^\star$ algorithm to attempt completing the action sequences that fail to change rack states to the goal pattern, thus saving incomplete yet valuable data for training. For the other, we use the A${}^\star$ algorithm to reduce redundancies in the collected training sequences. The reduction helps focus on actions that are more likely to contribute to the goal, thus making training more effective.

\section{Problem Definition and Solution Overview}
\label{sec:problem_definition_system_architecture}
This section first formulates the considered problem and then presents an overview of the proposed framework.

\subsection{In-Rack Test Tube Arrangement Problem}
We assume that a workspace contains a robot, a tube rack with test tubes, and other static obstacles. The rack comprises $n_r$ rows and $n_c$ columns of slots. There are $n_t$ distinct test tube types, and tubes of the same type are considered interchangeable. Since a slot may either be empty or occupied by a tube, the total number of possible arrangements of the rack is $(n_t+1)^{n_r n_c}$. These arrangements are represented as $\Gamma(\cdot)=\{\Lambda_1, \Lambda_2,\ldots\}$. A specific arrangement, $\Lambda_\mathrm{goal} \in \Gamma(\cdot)$, is designated as the goal pattern. The set of arrangements that meet the goal pattern is defined as $\Gamma(\Lambda_\mathrm{goal})  =\{\Lambda | \Lambda \subseteq \Gamma(\cdot)~\land~\Gamma(\Lambda) \subseteq \Gamma(\Lambda_\mathrm{goal})\}$. The set $\Gamma(\Lambda_\mathrm{goal})$ includes all arrangements where each test tube in it matches the type and position of those in $\Lambda_{goal}$. We define an action $\mathbf{a}_{jk\to pq}$ to be picking up a tube from the $j$-th row and $k$-column of the rack, and moving it to the $p$-th row and $q$-th column. Under these assumptions and definitions, we can formally define our problem as follows.

\begin{problem}[Multi-Class In-Rack Test Tube Rearrangement]
\textit{Given initial arrangements $\Lambda_\mathrm{init}$ for a set of test tubes, and a predefined goal pattern $\Lambda_\mathrm{goal}$, find a finite sequence of pick-and-place actions $(\mathbf{a}_1, \mathbf{a}_2, \ldots)$ that change the arrangements from $\Lambda_\mathrm{init}$ to an arrangement $\Lambda \in \Gamma(\Lambda_\mathrm{goal})$, while minimizing a cost metric $\Psi$.}
\end{problem}

\begin{figure}[!htbp]
  \begin{center}
  \includegraphics[width=\linewidth]{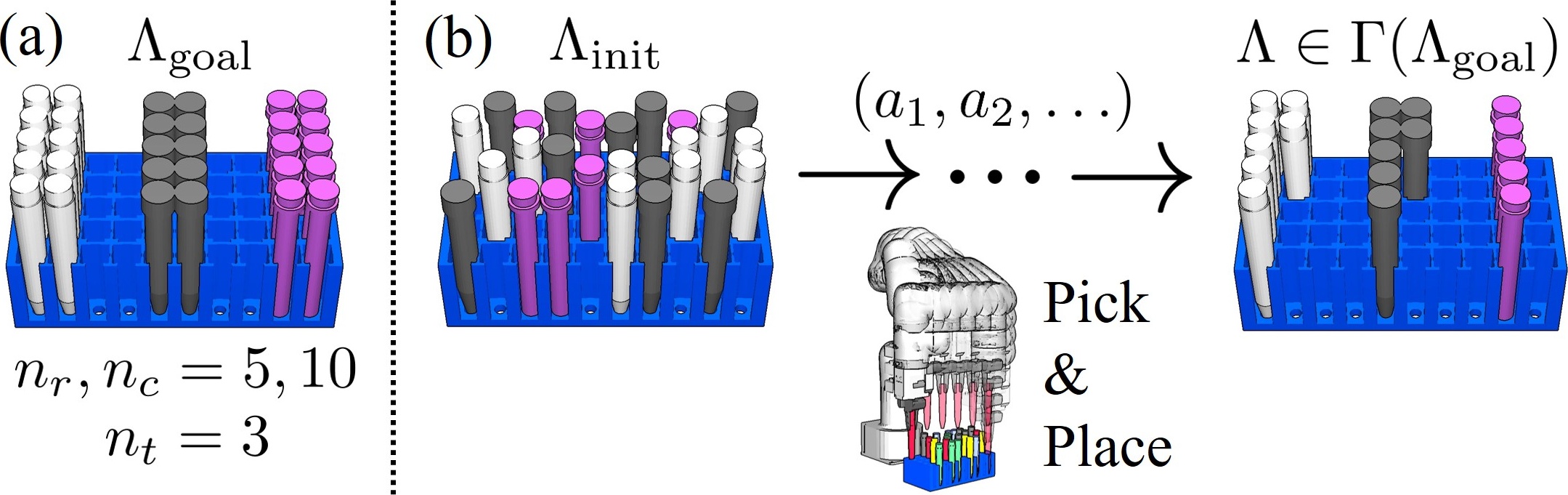}
  \caption{(a) An example of the goal pattern arrangement $\Lambda_\mathrm{goal}$ for a rack with $5\times10$ slots. $n_r$, $n_c$, $n_t$ denote the number of rows, columns and tube types, respectively. The types of test tubes are differentiated by color. (b) The objective of the multi-class in-rack test tube rearrangement problem is to transfer test tubes from an initial arrangement $\Lambda_\mathrm{init}$ to an target arrangement $\Lambda \in \Gamma(\Lambda_\mathrm{goal})$ using a minimal number of pick-and-place actions.}
  \label{fig:visual_demo1}
  \end{center}
\end{figure}

Here, the cost $\Psi$ could represent various metrics such as the number of actions, the robot's end-effector travel distance, etc. In this work, we focus solely on minimizing the number of pick-and-place actions. Fig. \ref{fig:visual_demo1} provides a visual example of an arrangement and the problem being discussed.

The multi-class in-rack test tube rearrangement problem presents unique challenges due to its combinatorial nature. The potential search space grows exponentially with the number of slots $n_r\times n_c$, and tube types $n_t$. Additionally, collision constraints introduce inter-dependencies between tubes. Moving one tube may require temporarily moving others out of the way. As the problem scales up, reasoning about these inter-dependencies and finding optimal solutions becomes intractable. It is extremely challenging to manually define heuristics that efficiently manage even a small rack. Therefore, we employ RL to solve this problem in this work. We aim to efficiently infer the sequence of actions online by using RL to explore and discover near-optimal solutions.

\subsection{Overview of the Proposed Framework} 
\label{sec:System Architecture} 
\begin{figure}[!htbp]
  \begin{center}
  \includegraphics[width=\linewidth]{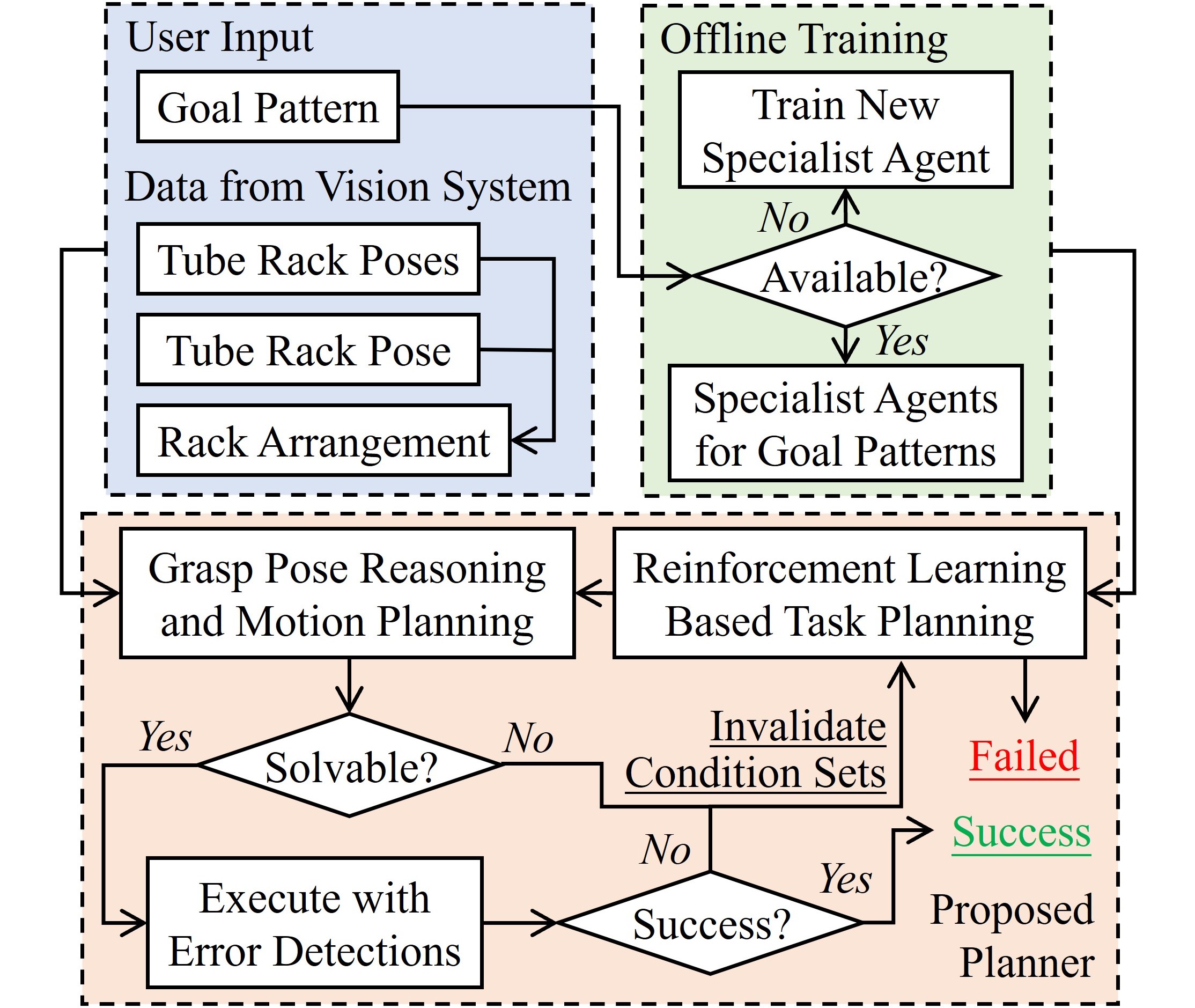}
  \caption{Flowchart of the proposed framework.}
  \label{fig:flowchart}
  \end{center}
\end{figure}

The workflow of the proposed framework is shown in Fig. \ref{fig:flowchart}. The upper-left dashed box denotes the input of the workflow. It includes:
\begin{itemize}
    \item \uline{Goal Pattern:} A predefined goal pattern, which represents the desired final arrangement of test tubes within the rack, is provided as input. A specialist agent is trained offline to learn policies for rearranging the test tubes into this goal pattern.
    \item \uline{Poses of Test Tubes and Tube Racks:} We detect the types and poses of test tubes and tube racks using RGB-D data. The details of the detection algorithm was published in our previous work \cite{chen2023rack}\cite{chen2023auto}. The detected results are accepted as input.
    \item \uline{Induced Rack Arrangement:} Based on the detected test tube poses and tube rack poses, we induce an initial rack arrangement. This initial rack arrangement is also accepted by the workflow as input. 
\end{itemize}

Considering the extensive variety of possible rack arrangements, each unique goal pattern is associated with a specialized RL agent for effective training. These specialist agents are trained offline using a distributed Q-learning structure. The upper-right dashed box of Fig. \ref{fig:flowchart} denotes the offline training process. The distributed Q-learning learns to generate action sequences for rearranging test tubes while considering potential collision constraints that may appear during grasping. If a goal pattern lacks a corresponding specialist agent, the system will train one prior to planning. 

The core of the framework is denoted by the dashed box at the bottom of Fig. \ref{fig:flowchart}. It includes two particular components. The ``Reinforcement Learning Based Task Planning" component incorporates the goal-specific specialist agent. It greedily uses the agent's policy to generate an action sequence and uses an A${}^\star$ trimmer to optimize the action sequence by removing redundant action steps. The component works at the task level. Its output is high-level commands like swapping tubes between slots in a grid world. Finger collisions are considered, but detailed kinematics and motion constraints are ignored. If the component fails to generate a feasible action sequence, the problem is considered to be unsolvable. 

Subsequently, the ``Grasp Pose Reasoning and Motion Planning" component processes the generated feasible action sequence to plan detailed pick-and-place motion. The component works at the lower motion level. It selects grasp poses according to task-level constraints and further checks their kinematic feasibility. In cases of failure, the module backtracks to the ``Reinforcement Learning Based Task Planning" component for re-generation and replanning while invalidating the maintained condition sets of the failed slots. The specialist agent policy would then select new actions considering constraints form by the invalidated conditions. When detailed pick-and-place motion is successfully planned, the workflow will move forward to robotic execution. However, environmental uncertainties may lead to execution failures. To mitigate this, the framework can incorporate various sensory feedback to monitor changes in execution. If inconsistency with planned results in the rack arrangement is detected, the workflow will backtrack to the same routine for re-generation and replanning.

The workflow is considered to be successfully completed when all tubes are rearranged according to the goal pattern. 

\section{Reinforcement Learning at the Task Level}
\label{sec:search_arrangement_sequence_using_rl}
The goal of RL on the task level is to determine which tubes to move while ignoring low-level robotic motion details. To this end, we will first introduce the Constrained Markov Decision Process (CMDP) formulation for our problem on the task level and then present the distributed Q-learning structure used in our work for efficient parallel training.

\subsection{CMDP Formulation}
Rearranging test tubes in a multi-class in-rack scenario can be represented as a Constrained Markov Decision Process (CMDP) \cite{altman2021constrained}. We employ a tuple $(\mathcal{S}, \mathcal{A}, \mathcal{P}, \mathcal{R}, \gamma, \mathcal{C})$ to define this process. In this tuple, $\mathcal{S}$ and $\mathcal{A}$ represent the sets of potential states and actions, respectively. $\mathcal{P}$ is a state transition model, which is deterministic in a specific problem context. $\mathcal{R}$ stands for the reward function of RL. $\gamma$ represents the discount factor. $\mathcal{C}$ represents the constraints limiting the action space for each state. Let $\pi_\mathcal{\theta}$ denote a neural policy $\pi_\mathcal{\theta}$ that maps from a state to an action while adhering to the constraints in $\mathcal{C}$. The subscript $\theta$ of the neural policy represents its parameter. The primary objective of the RL agent is to find a parameter $\theta^{\star}$ that maximizes the expected cumulative discounted return of the policy:
\begin{equation}
    \theta^{\star} = \underset{\theta}{\mathrm{argmax}}\underset{\mathcal{T}\sim\pi_\theta}{\mathbb{E}}
 [\sum_{t=0}^{\mathcal{H}}\gamma R(\mathbf{s}_t,\mathbf{a}_t,\mathbf{s}_{t+1})]\text{.}
\end{equation}
In this equation, $\mathcal{T}$ represents the action sequence selected based on $\pi_\theta$. $\mathcal{H}$ is the horizon or length of the sequence.

We design the state, action, constraints, reward, and termination criteria for the CMDP as follows:
 
\subsubsection{State}
\begin{figure}[!htbp]
  \begin{center}
  \includegraphics[width=\linewidth]{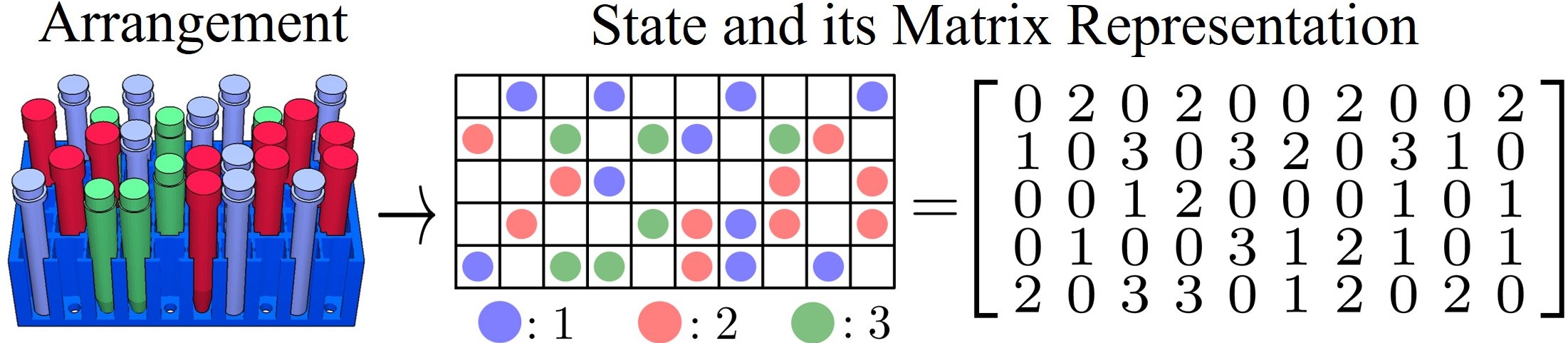}
  \caption{``Arrangement'' refers to the arrangement of test tubes in a rack. ``State'' is a ``Matrix Representation'' that encodes the rack arrangement.}
  \label{fig:state}
  \end{center}
\end{figure}

We define the rack arrangement $\Lambda$ as the state of our CDMP formulation and use $\mathbf{s}$ to denote it. Each $\mathbf{s}$ can be mathematically represented by a $n_r\times n_c$ matrix as follows:
\begin{equation}
    \mathbf{s}=
    \begin{bmatrix}
        e_{11} &
        e_{12} &
        \cdots &
        e_{1n_c}\\
        e_{21} &
        e_{22} &
        \cdots &
        e_{2n_c}\\
        \vdots & \vdots & \ddots & \vdots \\
        e_{n_r1} &
        e_{n_r2} &
        \cdots &
        e_{n_rn_c}
    \end{bmatrix}\text{.}
    \label{eq:vector_derivative}
\end{equation}
Here, each $e_{jk}$ indicates a slot at the $j$-th row and $k$-th column. Its value denotes the status of the slot. Specifically, $e_{jk}=0$ indicates an empty slot, while $e_{jk}=\lambda$ means that the slot contains a tube of type $\lambda\in\{1,\ldots n_t\}$. Fig. \ref{fig:state} illustrates an example of the state representation for an arrangement. 

Note that the goal patterns, which specify the target arrangement for test tubes, are not incorporated into the states. Embedding the goal patterns into a state would transform the RL challenge into a goal-conditioned problem \cite{liu2022goal}, significantly enlarging the search space and making convergence much harder for agents. Instead of incorporating the goal patterns into the states, we train individual specialist agents for each goal pattern to balance effective learning with practical training time considerations.

\subsubsection{Action} 
\begin{figure}[!htbp]
  \begin{center}
  \includegraphics[width=\linewidth]{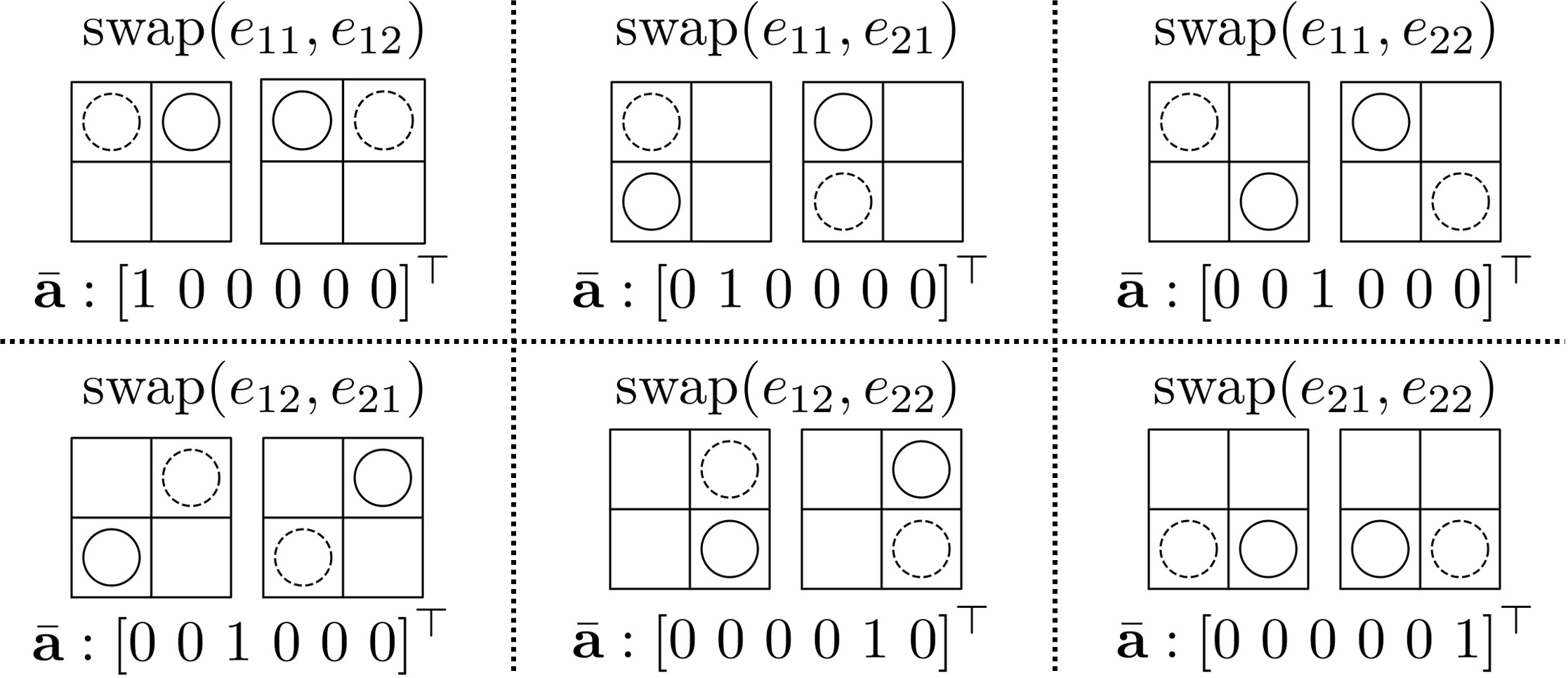}
  \caption{All potential actions for a rack with $2\times2$ slots. An action is defined as a swap between two slots without discerning the start and goal. Each distinct swap is uniquely encoded as a one-hot vector $\bar{\mathbf{a}}$ and can either represent moving a tube from a first slot to a second one or vice versa.}
  \label{fig:action}
  \end{center}
\end{figure}

We abstract the robotic pick-and-place action as a swap notation $\mathbf{a}:=\mathrm{swap}(e_{jk}, e_{pq})$ in our CMDP. It indicates exchanging the test tubes in slots $e_{jk}$ and $e_{pq}$. We assume during pick-and-place actions, only one test tube is moved and the robot can only manipulate one test tube at a time. Therefore, a swap action is valid if and only if one slot contains a test tube while the other is empty, and it is not permissible for a slot to swap with itself. Given a grid of slots consisting of $n_r$ rows and $n_c$ columns, the total number of feasible swap operations is $(n_rn_c)(n_rn_c-1)/2$. For computational representation, each swap action is encoded as a one-hot vector $\bar{\mathbf{a}} \in \mathbb{F}_2^{(n_rn_c)(n_rn_c-1)/2}$. Here, $\mathbb{F}_2$ indicates the finite field with two elements 0 and 1. One hot means there is a single ``1'' in the vector. The position of the single ``1'' distinctly identifies a specific swap operation. For example, considering a rack with $2\times2$ slots, all potential swap operations and their corresponding one-hot vector representations are illustrated in Fig. \ref{fig:action}. Representing actions as swaps effectively halves the action space compared to explicitly defining the pick-and-place actions, thereby improving training efficiency.

\subsubsection{Constraints}  
\begin{figure}[!htbp]
  \includegraphics[width=\linewidth]{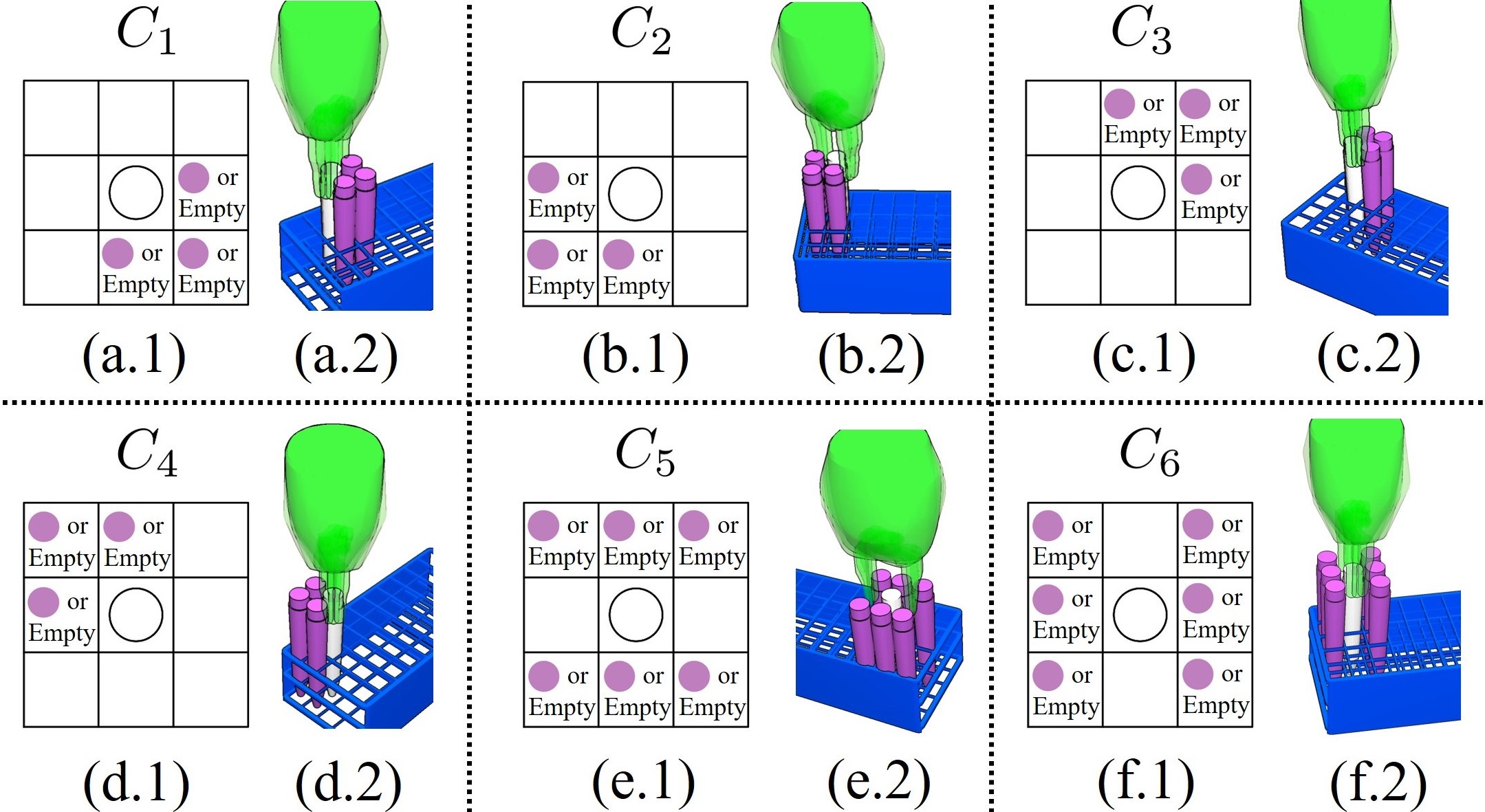}
  \caption{(a.1)-(f.1) are 6 different conditions. A test tube is considered acceptable by a gripper if one of these conditions is satisfied. The white circle in the center represents the tube that is going to be picked. The neighboring empty grids mean that the slot must be empty so that fingers can be positioned there without collision. There are no requirements on the grids with purple circles. They could either be filled with obstacle tubes or empty. (a.2)-(f.2) are corresponding collision-free grasp poses for each condition. The test tubes in pink are considered to be obstacle test tubes. And the test tube in white is the test tube going to be manipulated.}
  \label{fig:conditions}
\end{figure}

A swap action $\mathbf{a}:=\mathrm{swap}(e_{jk}, e_{pq})$ must meet certain constraints to prevent collisions and ensure feasibility. The constraints could include, for example, (i) one of the two slots being empty, and the other being occupied by a test tube; (ii) there must be enough space to pose the robotic fingers, etc. We formulate the constraints using the following condition equations.
\begin{subequations}\allowdisplaybreaks
\begin{align}
    &\begin{cases}
    \label{eq_slot_cond}
    ~C_1(x,y):\ E_1 + e_{x-1, y} + e_{x-1, y+1} = 0\\
    ~C_2(x,y):\ E_2 + e_{x-1, y} + e_{x-1, y-1} = 0\\
    ~C_3(x,y):\ E_1 + e_{x+1, y} + e_{x+1, y+1} = 0\\
    ~C_4(x,y):\ E_2 + e_{x+1, y} + e_{x+1, y-1} = 0\\
    ~C_5(x,y):\ e_{x, y-1} + e_{x, y+1} = 0\\
    ~C_6(x,y):\ e_{x-1, y} + e_{x+1, y} = 0\\
    \end{cases},\\
    & ~~~~~~ \left(
                \begin{array}{ll}
                E_1 = e_{x-1, y-1} + e_{x, y-1} + e_{x+1, y-1} \\
                E_2 = e_{x-1, y+1} + e_{x, y+1} + e_{x+1, y+1}\\
                \end{array}\right)\nonumber\\ 
    &C_7:(e_{jk}= 0\land e_{pq}>0) \lor (e_{jk} > 0\land e_{pq}=0)\text{.} 
\end{align}
\end{subequations}
Here, the values $(x,y)$ represent the coordinates of a possible tube location (slot). It could be replaced by either $(j,k)$ or $(p,q)$ involved in the swap. The six conditions in \eqref{eq_slot_cond} respectively ensure that there is enough space for positioning the robotic fingers and achieving collision-free grasping. Fig. \ref{fig:conditions} visually depicts the free spaces guaranteed by these conditions. The condition in $C_7$ ensures that the swap action is valid by requiring one slot to contain a test tube while the other slot to be empty. 

Based on this formulation, robot fingers can be posed a round a tube at $(x,y)$ without collision if any of the six equations in \eqref{eq_slot_cond} is met:
\begin{equation}
    \bigvee_{w=1\sim6} C_{w}(x,y),
\end{equation}
and a swap action is defined as acceptable if the following logical constraint is true:
\begin{equation}
     \mathrm{acc}(\mathbf{a}): (\bigvee_{w=1\sim6} C_{w}(j,k)) \land (\bigvee_{w=1\sim6} C_{w}(p,q)) \land C_7,
     \label{eq_cond_a_all}
\end{equation}
For a state $\mathbf{s}\in \mathcal{S}$, its set of acceptable actions is defined as
\begin{equation}
\mathcal{A}_\mathrm{acc}(\mathbf{s}) = \{\mathbf{a}\ |\ \mathrm{acc}(\mathbf{a})>0\}.
\label{eq_af}
\end{equation}
The proposed framework only explores the actions in $\mathcal{A}_\mathrm{acc}(\mathbf{s})$ when training the RL model. This limited exploration helps reduce the search space when collecting the training data, thus improving training efficiency and operational reliability. 

Note that the condition equations only consider collisions at the task level. They ensure robotic fingers can be posed around a tube in a slot without collision, but do not take into account other constraints like robot kinematics, workspace obstacles, etc. These other constraints will be solved at the motion level and will be fed back to the task level by maintaining a condition set for each rack slot. We will discuss more about motion planning and motion-level feedback when presenting the inference model in Section V.C and motion generation and replanning in Section VI.C.

\subsubsection{Reward} 
\begin{figure}[!htbp]
  \begin{center}
  \includegraphics[width=\linewidth]{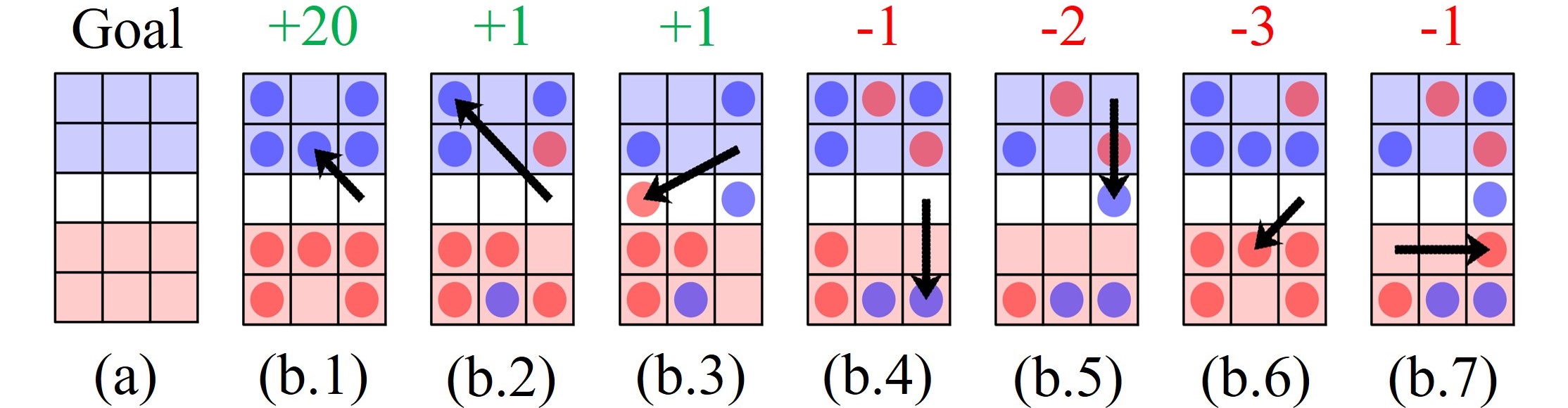}
  \caption{(a) Goal pattern. Blue tubes are expected to be moved tot he blue cells. Red tubes are expected to be moved to the red cells. The white cells are expected to be kept empty. (b.1)-(b.7) Exemplary cases when the reward function $r(\mathbf{s}_{t}, \mathbf{a}_{t}, \mathbf{s}_{t+1})$ have different values.}
  \label{fig:reward}
  \end{center}
\end{figure}

Particularly, we considered the following cases when designing the reward function $r(\mathbf{s}_{t}, \mathbf{a}_{t}, \mathbf{s}_{t+1})$, and handcrafted the reward function values by experience. Table \ref{tab:reward_function} shows a summary of the cases and the designed reward values. 
\begin{itemize}
    \item \uline{Goal Achievement:} The rearrangement task is considered successful when all test tubes are correctly positioned within the goal pattern. Fig. \ref{fig:reward}(b.1) exemplifies a goal achievement. We let the reward function return a high value ($+20$) when the goal is achieved.
    \item \uline{Place a Tube in a Correct Goal Slot:} When a tube is moved from a non-goal pattern slot to a position within the goal pattern, we let the reward function return a moderate value ($+1$). This reward setting helps encourage actions that progress toward the goal pattern. Fig. \ref{fig:reward}(b.2) shows an example of the case.
    \item \uline{Remove a Tube from an Incorrect Goal Slot:} We also let the reward function return a moderate value ($+1$) when a tube is moved out of an incorrect goal pattern slot. This reward setting will also encourage actions that progress toward the goal pattern. Fig. \ref{fig:reward}(b.3) shows an example of this case.
    \item \uline{Place a Tube in an Incorrect Goal Slot:} The reward function returns $-1$ when a tube is moved from a non-goal pattern slot to an incorrect goal pattern slot. This reward setting helps discourage counterproductive actions. Fig. \ref{fig:reward}(b.4) illustrates an example. 
    \item \uline{Remove a Tube from a Correct Goal Slot:} When a tube is moved from a goal pattern slot to a non-goal pattern location, we let the reward function return $-2$. This helps discourage actions that undermine the task’s objectives. Fig. \ref{fig:reward}(b.5) shows an example of this case.
    \item \uline{Block Available Goal Slots:} The reward function is set to return $-3$ if a tube is moved to a goal pattern slot but obstructs accessible empty slots needed for remaining tubes. Fig. \ref{fig:reward}(b.6) exemplifies such as case.
    \item \uline{Reach Deadlock:} A substantial penalty $-20$ is applied to actions that lead to a deadlock where no further progress is possible. This strongly discourages actions that trap the agent in an infeasible situation.\footnote{Deadlocks are rare. Reaching a deadlock will lead to a fatal failure.}  
    \item \uline{Indifferent Movements:} Actions other than the above cases are non-progressive or redundant. They are considered indifferent and a penalty reward ($-1$) is applied to them to avoid long and useless action sequences. Fig. \ref{fig:reward}(b.7) shows an example of this case.
\end{itemize}

\begin{table}[!h]
\centering
\caption{\label{tab:reward_function}Handcrafted Rewards}
\begin{tabularx}{\linewidth}{>{}l|>{\leavevmode}X}
\toprule
 Case & Reward Value\\
\midrule 
Goal achievement &  $+20$\\
Place a tube in a correct goal slot & $+1$  \\
Remove a tube from an incorrect goal slot  & $+1$  \\
Place a tube in an incorrect goal slot   & $-1$  \\
Remove a tube from a correct goal slot  & $-2$  \\
Block available goal slots  & $-3$  \\
Reach deadlock  & $-20$  \\
Other movements  & $-1$  \\
\bottomrule
\end{tabularx}
\end{table} 

\subsubsection{Termination Criteria} 
There are three distinct termination criteria.
\begin{itemize}
    \item \uline{Goal Achievement:} Termination may occur when the goal pattern is achieved (the rearrangement task is successfully completed). This is determined by matching the locations of test tubes in $\mathbf{s}$ with those in the goal pattern. The rearrangement task is considered completed when all test tubes are correctly positioned within the goal pattern. 
    \item \uline{Deadlock:} Termination may occur when the rack reaches a deadlock where no further valid actions exist.
    \item \uline{Horizon Limit:} Termination occurs when the sequence length reaches a predefined horizon limit $\mathcal{H}$. This ensures the process does not continue infinitely. 
\end{itemize}

\subsection{Distributed Q-learning}
\begin{figure}[!htbp]
  \begin{center}
  \includegraphics[width=\linewidth]{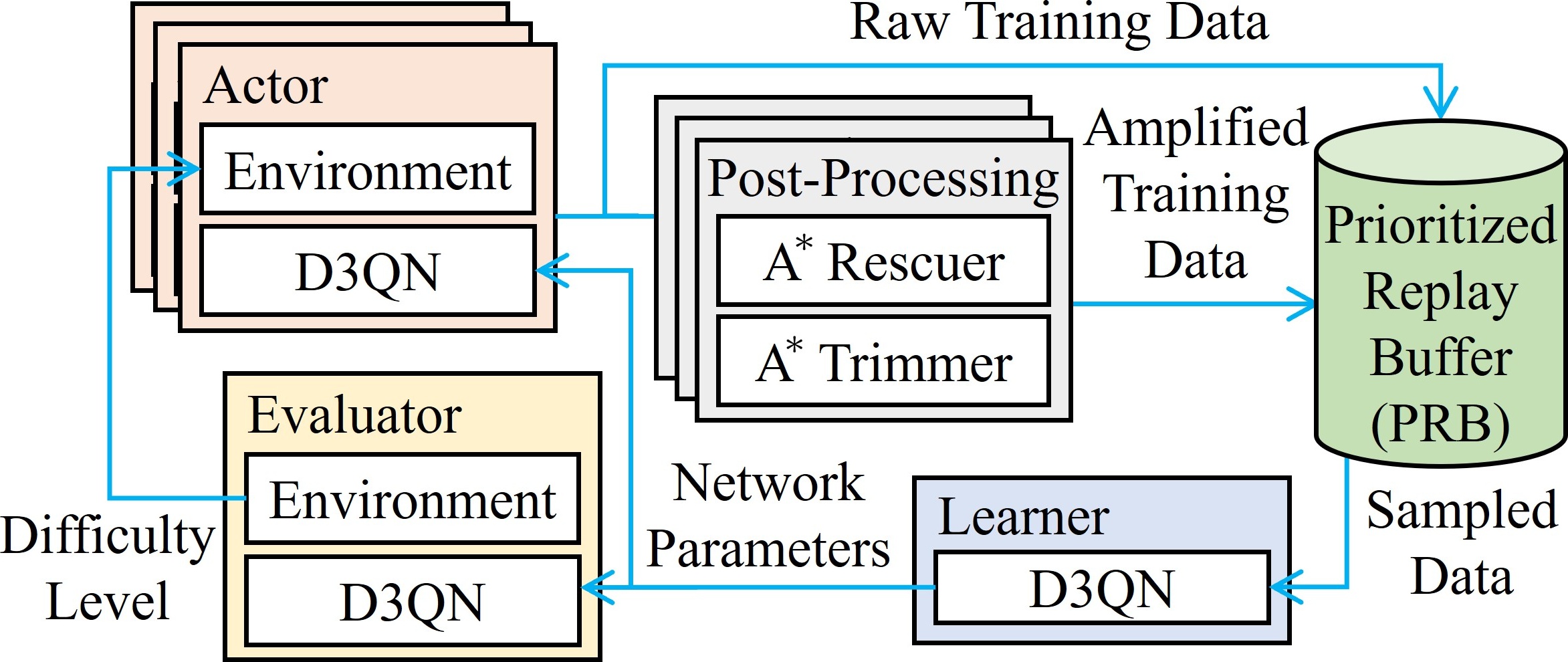}
  \caption{Flowchart of the proposed distributed deep Q-learning structure.}
  \label{fig:distributed_q_learning}
  \end{center}
\end{figure}

We implement a distributed deep Q-learning structure based on ApeX DQN \cite{horgan2018distributed} to improve search efficiency. The distributed deep Q-learning retains the original ApeX DQN's feature of multiple actors, centralized learner, and centralized prioritized replay buffer. Meanwhile, it replaces the DQN with D3QN (Dueling Double Deep Q Network) \cite{van2016deep, wang2016dueling}, employs an additional post-processing technique for amplifying the explored data and an evaluator module for curricularly regulating environmental difficulty. The diagram in Fig. \ref{fig:distributed_q_learning} shows the components and workflow of the distributed deep Q-learning structure. The details of each component in the structure are described below, and the details of the D3QN network is presented in the appendix: 
\begin{itemize}
    \item \uline{Actors:} There are multiple actors to interact with various environmental instances for fast and diverse data collection. Each agent employs an $\epsilon$-greedy strategy for environmental interaction. This parallel approach accelerates the learning process and facilitates the exploration of state space. The parameters of the agent's D3QN network are periodically updated according to the parameters of a centralized learner.
    \item \uline{Learner:} The learner is a centralized module. It continuously updates the D3QN network based on data provided by the actors. The updating is grounded in the loss function defined in equation \eqref{eq:q_network_update} and incorporates a soft strategy that gradually blends the weights of the target network with the weights of the primary network. Periodically, the learner synchronizes its primary neural network parameters with the actors and the evaluator.
    \item \uline{Prioritized Replay Buffer (PRB):} The PRB stores and manages transitions (states, actions, rewards, next states, and termination flags) collated from the actors. The PRB continuously updates transition priorities based on the magnitude of their loss during training to ensure that the learner focuses on the most challenging or unexpected experiences \cite{schaul2015prioritized}. During sampling, the PRB biases towards transitions with higher priorities to accelerate the overall learning efficiency. 
    \item \uline{Evaluator for Curriculum Learning:} The evaluator module assesses the agent's performance and decides when to increase the complexity of tasks, thereby progressively guiding the agent to go through more challenging scenarios. This progressive increase is essential for effective learning, as it prevents the agent from getting overwhelmed by complex tasks at early stages and ensures that foundational skills can be developed before tackling more advanced challenges. A more detailed elaboration on this process can be found in Section \ref{sec:curriculum_learning}.
    \item \uline{Post-Processing with A${}^\star$ Rescuer and A${}^\star$ Trimmer:} 
    We implement an A${}^\star$ rescuer and an A${}^\star$ trimmer based on the A${}^\star$ algorithm presented in our previous work \cite{wan2022arranging} to amplify the collected training data and thus accelerate convergence. The A${}^\star$ rescuer attempts to complete the action sequences that fail to transfer rack states to the goal pattern. It helps to save incomplete yet valuable data for training. The A${}^\star$ trimmer attempts to reduce redundant action sequences and relabel rewards accordingly. It helps focus exploration on actions that more likely contribute to the goal. In the following section, we present the algorithmic details of the rescuers and trimmers and discuss their role in advancing the framework.
\end{itemize} 

\section{Post-Processing, Training, and Inference}

In this section, we first detail our A${}^\star$-based post-processing technique used to amplify the collected training data. Then, we discuss additional strategies used during training, including curriculum learning and tabu search. Finally, we present our inference methods for the application phase.

\subsection{A${}^\star$-Based Post-Processing}
We implement an A${}^\star$ rescuer and an A${}^\star$ trimmer to amplify the collected training data and thus accelerate convergence. 

\subsubsection{A${}^\star$ Rescuer}
The reason we included an A${}^\star$ rescuer is that the horizon limitations sometimes interrupt the reinforcement exploration and lead to incomplete data. Previously, the interrupted data was ignored, which made data collection and training significantly slow. In this work, we employ an A${}^\star$ rescuer to attempt the completion of action sequences that fail to transfer rack states to the goal pattern. The rescuer uses an A${}^\star$ algorithm to connect an interrupted state to a goal state. If the connection is successful, a new sequence is obtained and undergoes additional training. The process enables the rescuer to save incomplete yet valuable data for training and thereby improves learning performance. 

\begin{figure}[!htbp]
  \begin{center}
  \includegraphics[width=\linewidth]{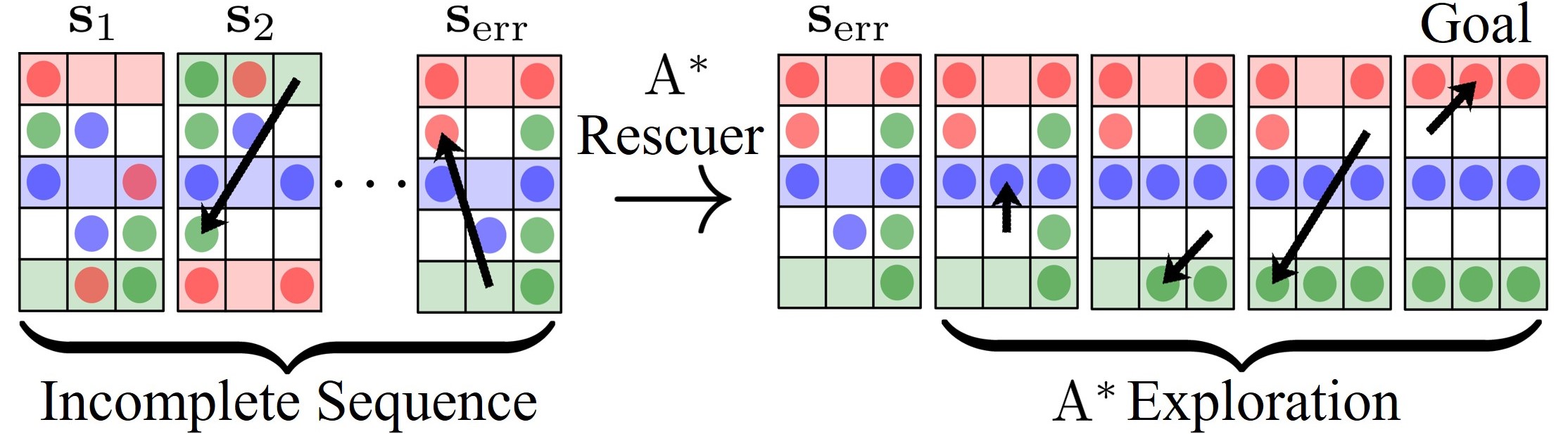}
  \caption{(a) The A${}^\star$ rescuer attempts to complete sequences terminated by horizon limits. It employs an A${}^\star$ algorithm \cite{wan2022arranging} to connect the interrupted state $\mathbf{s}_\mathrm{err}$ to a state that meets the goal pattern. (b) The computation of $g(\cdot)$ and $h(\cdot)$ in the A${}^\star$ algorithm for node expansion.}
  \label{fig:A_rescuer}
  \end{center}
\end{figure}

We present the algorithmic details of the A${}^\star$ rescuer using the example shown in Fig. \ref{fig:A_rescuer}. The left part of the figure shows an incomplete state sequence $(\mathbf{s}_1, \ldots, \mathbf{s}_\mathrm{err})$ that was terminated at state $\mathbf{s}_\mathrm{err}$. The A${}^\star$ rescuer utilize the interrupted state $\mathbf{s}_\mathrm{err}$ as the initial node for the the A${}^\star$ algorithm \cite{wan2022arranging}. It selects the next node based on the lowest $g(\cdot) + h(\cdot)$, where $g(\cdot)$ (the cost from the initial node to a candidate node) is computed as the number of actions, and $h(\cdot)$ (the heuristic estimation) is computed as the number of tubes that are not in the designated slots in the goal pattern. They are mathematically represented as: 
\begin{equation}
\begin{cases}
    ~g(\mathbf{s}_i) = i\\ 
    ~h(\mathbf{s}_i, \mathbf{s}_g) = \sum_{p}^{n_r}\sum_{q}^{n_c} (\mathbbm{1}( (e^{\mathbf{s}_i}_{pq} >0) \land (e^{\mathbf{s}_i}_{pq} \neq e^{\mathbf{s}_g}_{pq})))
\end{cases},
\end{equation}
where $\mathbf{s}_g$ is the input goal state of the A${}^\star$ algorithm and $\mathbf{s}_i$ is a state to be expanded. The $e^{\mathbf{s}}_{pq}$ represents the status of the slot at the $p$-th row and $q$-th column of state $\mathbf{s}$. $\mathbbm{1}(\cdot)$ is an indicator function that has a value of $1$ when conditions in $(\cdot)$ are met. The value is $0$ when the conditions are not satisfied. 

The A${}^\star$ searches for a path to a goal state within a specified iteration limit. If a solution is found, the states along the path are converted into transitions and are added to the replay buffer. Otherwise, if no solution is found within the iteration limit, the interrupted state $s_\mathrm{err}$ is considered too distant from any potential goal states, and the effort for completing the interrupted state sequence is abandoned.

\subsubsection{A${}^\star$ Trimmer} 
\label{sec:A_star_relabeler}
On the other hand, successfully collected training data may exhibit considerable redundancy. This is because the test tube rearrangement problem has a large state and action space. The rewards are relatively sparse and an actor might repeatedly attempt ineffective or irrelevant actions without realizing that they are meaningless for achieving the goal. To avoid including meaningless actions for training, we employ an A${}^\star$ trimmer to reduce the collected training data. The reduction helps focus on actions that are more likely to contribute to the goal, thus making training more effective.

Consider a complete sequence collected by an actor. Essentially, the sequence comprises a state sequence $(\mathbf{s}_1, \ldots, \mathbf{s}_n)$ and their corresponding action sequence $(\mathbf{a}_1,\ldots, \mathbf{a}_{n-1})$ and reward sequence. We consider a sequence to be potentially redundant when a sequence of actions connecting two states can be substituted by a shorter one. The following equation shows the potential redundancies mathematically:
\begin{align}
    &\exists\mathbf{s}_i,~\mathbf{s}_j,~\exists(\mathbf{a}^\prime_1, \ldots, \mathbf{a}^\prime_{T}), \nonumber\\   
    &~~~~\begin{cases}
    \mathbf{s}_i,~\mathbf{s}_j\in(\mathbf{s}_1, \ldots, \mathbf{s}_n)\\
    \mathbf{s}_i\xrightarrow{(\mathbf{a}^\prime_1\ldots\mathbf{a}^\prime_{T})}\mathbf{s}_j\\
    T<(j-i-1)\\
    \forall\mathbf{a}^\prime\in(\mathbf{a}^\prime_1, \ldots, \mathbf{a}^\prime_{T}), \  \mathrm{acc}(\mathbf{a'})>0.
    \end{cases}
\end{align}\allowdisplaybreaks 
Here, $j-i-1$ is the length of the action sequence corresponding to the state sequence $(\mathbf{s}_i$, \ldots, $\mathbf{s}_j)$. $(\mathbf{a}^\prime_1, \ldots. \mathbf{a}^\prime_{T})$ represents a shorter action sequence with length $T$ that can transit from state $\mathbf{s}_i$ to $\mathbf{s}_j$. 

To quickly identify potential redundancies, we define a function ``$\mathrm{disorder(\cdot)}$'' to measure the difference between two states:
\begin{equation}
    \mathrm{disorder(\mathbf{s}_i, \mathbf{s}_j)} =  
    \sum_{p}^{n_r}\sum_{q}^{n_c} (\mathbbm{1}(e^{\mathbf{s}_i}_{pq} >0) - 
    \mathbbm{1} (e^{\mathbf{s}_i}_{pq} = e^{\mathbf{s}_j}_{pq} > 0)),
\end{equation}\allowdisplaybreaks 
where the $e^{\mathbf{s}}_{pq}$ represents the status of the slot at the $p$-th row and $q$-th column of state $\mathbf{s}$. $\mathbbm{1}(\cdot)$ is an indicator function that has a value of $1$ when conditions in $(\cdot)$ are met. The value is $0$ when the conditions are not satisfied. The ``$\mathrm{disorder(\cdot)}$'' essentially counts the number of tubes that are not aligned between two states. Since a non-redundant action increases the disorder by $1$, redundancies may occur if $\mathrm{disorder(\mathbf{s}_i, \mathbf{s}_j)}$ is smaller than the number of actions between $\mathbf{a}_i$ and $\mathbf{a}_j$, namely $\mathrm{disorder(\mathbf{s}_i, \mathbf{s}_j)} < j-i-1$.

\begin{figure}[!htbp]
  \begin{center}
  \includegraphics[width=\linewidth]{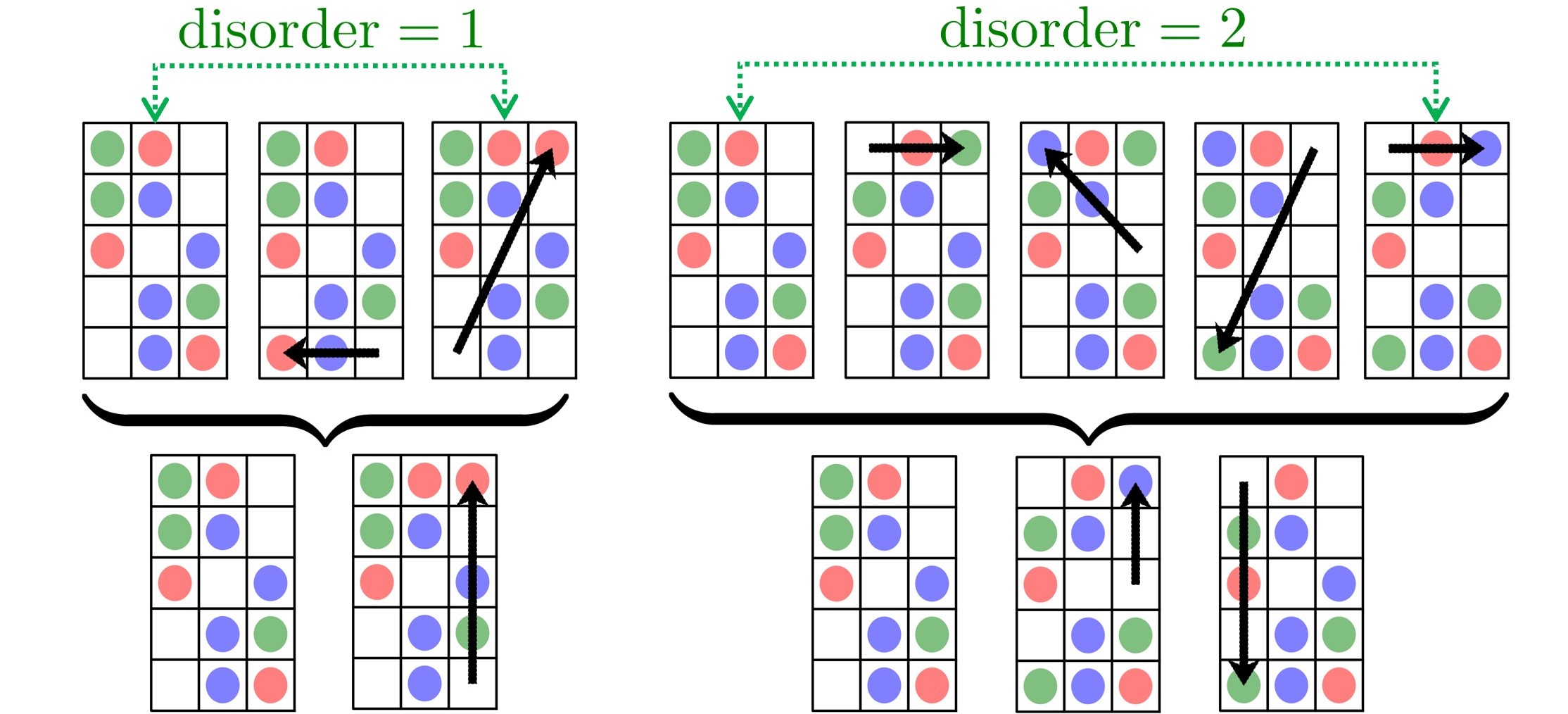}
  \caption{Exemplary sequences with potential redundancies. The first row is the state sequence with potential redundancies. The second row is the state sequences reduced by the A${}^\star$ trimmer.}
  \label{fig:refined_path}
  \end{center}
\end{figure}

Fig \ref{fig:refined_path} illustrates two examples of potential redundancies, their disorder values, as well as the results after eliminating these redundancies. The A${}^\star$ trimmer attempts to reduce the potential redundancies by using the same A${}^\star$ algorithm as the A${}^\star$ rescuer. It searches for a path between the start state and end state of a redundant sequence using the A${}^\star$ algorithm with a predefined limit. If a shorter action sequence can be found, the A${}^\star$ trimmer will replace the original sequence section and thus reduce redundancies. Otherwise, if no solution can be found or the solution length is equal to the original sequence section length, the A${}^\star$ trimmer ignores the results.

Note that action sequences that involve temporary removal of inter-tube obstructions, i.e. temporarily moving a tube outside the goal pattern to allow other tubes to move, may also be recognized as having potential redundancies. Such potential redundancies are not really redundant and are globally necessary. The A${}^\star$ trimmer will have a high probability of returning the same sequence in the presence of such redundancies.

\begin{figure}[!htbp]
  \begin{center}
  \includegraphics[width=\linewidth]{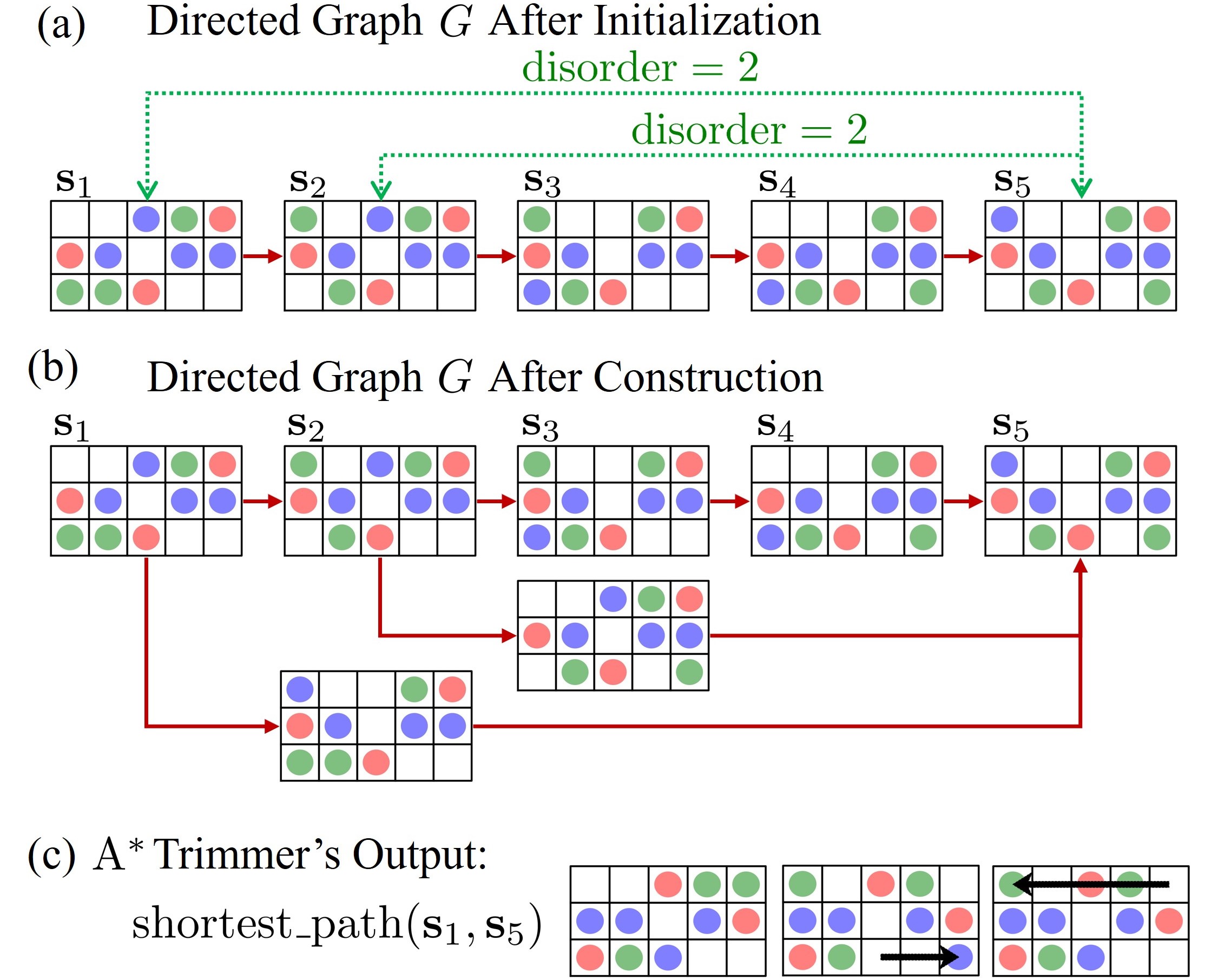}
  \caption{(a)Directed graph $G$ after initialization for state sequence $(s_1, \ldots, s_5)$. The red arrow represents a edge with weight $1$ (2) Directed graph $G$ after construction. (c) The A${}^\star$ trimmer reduces the state sequence $(s_1, \ldots, s_5)$ by finding the shortest path in $G$.}
  \label{fig:directed_graph}
  \end{center}
\end{figure}

Algorithm \ref{alg:remove_redundant_actions} shows implementation details of the trimmer. The input to the algorithm includes a complete state sequence $S$ returned by an actor, an iteration limit $L$ for the A${}^\star$ algorithm, and a horizon limit $H_\mathrm{trm}$ for the trimming process. The $L$ limit trades off computation time with successful redundancy removal rates, as more iterations for A${}^\star$ leads to higher success in searching a path between two states. The $H_\mathrm{trm}$ sets a threshold for assessing redundancies between states: If the disorder value between two states exceeds $H$, it is assumed that the redundancies are unlikely to be trimmed by A${}^\star$ algorithm within $L$ iterations. The algorithm's output is a trimmed state sequence. \footnote{Note that the algorithm cannot remove all redundancies. This is not negative as keeping some redundancies helps avoid sparse negative rewards.}

The essential section of the algorithm constructs a directed graph based on the sequence $S$. This is because the redundancies might be nested along $S$. Simply applying A${}^\star$ search by iteration may not lead to a globally shortest path. Fig. \ref{fig:directed_graph}(b) illustrates an example of the nested redundancies. To globally shorten the nested redundancies, the algorithm scans all potential two states of $S$ to compute their disorders and identify the redundancies (lines 6-8, Algorithm \ref{alg:remove_redundant_actions}). If two states can be connected by a shorter sequence section, they will be connected by directed edges (lines 9-24, Algorithm \ref{alg:remove_redundant_actions}). The algorithm constructs a directed graph after the connection. It then employs conventional graph search methods to identify the shortest state sequence $S^{^\prime}$ from $G$ (line 25, Algorithm \ref{alg:remove_redundant_actions}).\footnote{NetworkX (\url{https://networkx.org/}) is utilized for graph operations.}

\begin{algorithm}[!htbp]
\SetAlgoLined
\SetAlFnt{\footnotesize}
\DontPrintSemicolon
\newcommand\mycommfont[1]{\footnotesize\ttfamily\textcolor{blue}{#1}} 
\SetCommentSty{mycommfont}
\SetKwData{True}{True}
\SetKwFunction{DirectedGraph}{DirectedGraph}
\SetKwFunction{add}{add}
\SetKwFunction{addNode}{addNode}
\SetKwFunction{addEdge}{addEdge} 
\SetKwFunction{shortestPath}{shortestPath} 
\SetKwFunction{Astar}{A${}^\star$\_Search} 
\SetKwFunction{Disorder}{disorder} 
\SetKwFunction{existAction}{existAction} 
\SetKwFunction{listTuple}{list} 
\KwIn{State sequence $(\mathbf{s}_1, \mathbf{s}_2, ..., \mathbf{s}_n)$; Iteration limit $L$ for the A${}^\star$; Horizon limit for trimming $H_\mathrm{trm}$}
\KwOut{A trimmed state sequence}
    $G\leftarrow$\DirectedGraph{}\\
    \For{$i~\textnormal{in}~(1,2,...,n)$}{
        $G.$\addNode{$\mathbf{s}_i$}\\
        \uIf{$i>1$\ $\wedge$\ $s_{i-1}\neq s_i$}{
            $G.$\addEdge{$\mathbf{s}_{i-1}, \mathbf{s}_i, 1$}
        }
    }
    \For{$i~\textnormal{in}~(1,2, ..., n-2)$}{
        \For{$j~\textnormal{in}~(n,n-1,...,i+2)$}{ 
            $d\leftarrow$\Disorder{$\mathbf{s}_i,\mathbf{s}_j$} \\
            \uIf{$d > 1$}{ 
                \uIf{$d < j- i < H_\mathrm{trm}$}{
                    $S_\mathrm{tmp} \leftarrow$\Astar{$s_i, s_j, L$}\\
                    \For{$k~\textnormal{in}~(1,2,...,|S_\mathrm{path}|)$}{
                            \uIf{$S_\mathrm{tmp}[k]\notin G$}{
                                $G.$\addNode{$S_\mathrm{tmp}[k]$}\\
                            }
                            \uIf{$k>1\ \wedge\ S_\mathrm{tmp}[k-1]\neq S_\mathrm{tmp}[k]$}{
                                $G.$\addEdge{$S_\mathrm{tmp}[k-1], S_\mathrm{tmp}[k], 1$}\\
                            }
                        }
                    \textbf{break}\\
                }
            }
            \uElse{
                \uIf{$s_i==s_j$}{
                    $G.$\addEdge{$\mathbf{s}_i, \mathbf{s}_j, 0$}\\
                    \textbf{break}
                }
                \uElseIf{\existAction{$s_i,s_j$}}{ 
                        $G.$\addEdge{$\mathbf{s}_i, \mathbf{s}_j, 1$}\\
                        \textbf{break}
                }
            }
        }
    }
    \textbf{return} $G.$\shortestPath{$\mathbf{s}_1, \mathbf{s}_n$}\\   
 \caption{A${}^\star$ Trimmer}
 \algorithmfootnote{
     \begin{description}
         \item \texttt{DirectedGraph()} Create a directed graph.
         \item \texttt{addNode()} Add a node to a directed graph.
         \item \texttt{addEdge()} Add an edge between two given nodes in a directed graph. The last parameter indicates the weight of the edge.
         \item \texttt{shortestPath()} Find the shortest path between two nodes in a directed graph using Dijkstra's algorithm.
         \item \texttt{A${}^\star$\_search()} Find the shortest path between two states using the A${}^\star$ search algorithm. It returns an empty sequence if a path cannot be found within a specified iteration limit.
         \item \texttt{existAction()} Examine if accessible actions exist between two states.
    \end{description}
 }
\label{alg:remove_redundant_actions}
\end{algorithm}

\subsection{Other Learning Strategies}
Both the data collected by the actors and the data post-processed by the A${}^\star$ rescuer and trimmer are included in the PRB for training. During training, we included several learning strategies like curriculum learning and tabu search to improve efficiency and expedite convergence. Their details will be presented in this subsection.

\subsubsection{Curriculum Learning}
\label{sec:curriculum_learning}
We employ a curriculum learning approach to evolve the agent's training problem setting from simple to difficult. The approach can help expedite convergence and improve task performance \cite{narvekar2020curriculum}. The algorithmic implementation of curriculum learning used in our framework is outlined in Algorithm \ref{alg:cl}.

In particular, we define two parameters ($\sigma_\mathrm{tubes}$, $\sigma_\mathrm{displaced}$) to regulate the difficulty of the problem used for training. $\sigma_\mathrm{tubes}$ regulates the number of test tubes in the problem. $\sigma_\mathrm{displaced}$ regulates the number of test tubes not at their goal slots. Larger values of these parameters lead to a more challenging problem. 
Initially, the agent experiences a problem with low values of $\sigma_\mathrm{tubes}$ and $\sigma_\mathrm{displaced}$ (line 1, Algorithm \ref{alg:cl}). As the training progresses, the agent's performance is periodically evaluated over $N$ rearrangement trials (lines 4-14, Algorithm \ref{alg:cl}). If the agent successfully solves at least $N_\mathrm{thres}$ trials, the difficulty increases (lines 8-14, Algorithm \ref{alg:cl}). Note that in the particular flow, we increase $\sigma_\mathrm{displaced}$ first while maintaining $\sigma_\mathrm{tubes}$ constant. This ensures that the agent develops a robust foundation in addressing diverse scenarios within a fixed $\sigma_\mathrm{tubes}$. Once the agent successfully learns to handle all possible displacements of test tubes for a given $\sigma_\mathrm{tubes}$, we proceed to increase $\sigma_\mathrm{tubes}$ (lines 10-12, Algorithm \ref{alg:cl}). The parameters and difficulty adjustment repeats until $\sigma_\mathrm{tubes}$ exceeds a predefined threshold $\sigma_\mathrm{tubes}^\star$. The algorithm will output an optimal D3QN parameters $\theta^\star$ when stopped (lines 13, Algorithm \ref{alg:cl}).

In addition to the complexity adjustment, we evaluate the correctness of the solutions for simple problems to secure effective training in the early stage. The performance length threshold parameter $N_\mathrm{reg}$ used at line 6 of Algorithm \ref{alg:cl} regulates the length of the early stage. In the evaluation, we regard a solution as correct only if its length is less than or equal to the number of test tubes. The evaluation ensures that the agent learns to solve tasks efficiently before advancing. However, the evaluation is only applicable to simple problems in the early stage. As the problems become more difficult, inter-dependencies among test tubes grow, and longer action sequences become necessary to solve the tasks successfully. We thus stop evaluating the solution lengths.


\begin{algorithm}[!htbp]
\SetAlgoLined
\SetAlFnt{\footnotesize}
\DontPrintSemicolon
\SetKwData{True}{True}  
\SetKw{Break}{break}
\SetKwFunction{setDifficulty}{setDifficulty}
\SetKwFunction{synchronize}{synchronize}
\SetKwFunction{evalPerformance}{evalPerformance}
\KwIn{A problem $P$; Performance threshold $N_\mathrm{thres}$; Evaluation interval $I_\mathrm{eval}$; Maximum test tube number $\sigma_\mathrm{tubes}^\star$; A shared D3QN agent $\pi_{\theta}$ with a parameters $\theta$; Horizon limit $\mathcal{H}$; Performance length threshold $H_\mathrm{var}$}
\KwOut{Optimal D3QN parameters}
    $\sigma_\mathrm{tubes},\ \sigma_\mathrm{displaced}\leftarrow 1$\\  
    $P.$\setDifficulty{$\sigma_\mathrm{tubes}$, $\sigma_\mathrm{displaced}$}\\ 
    \While{True}{
        \synchronize{$\theta, I_\mathrm{eval}$}\\ 
        $H_\mathrm{var}\leftarrow\mathcal{H}$\\ 
        \lIf{$\sigma_\mathrm{tubes}<N_\mathrm{reg}$}{   
            $H_\mathrm{var}\leftarrow\sigma_\mathrm{tubes}$
        }  
        $N_\mathrm{success}\leftarrow$\evalPerformance{$\pi_{\theta}$, $P$, $H_\mathrm{var}$}\\
        \If{$N_\mathrm{success}\geq N_\mathrm{thres}$}{ 
            $\sigma_\mathrm{displaced} \leftarrow \sigma_\mathrm{displaced} +1$ \\
            \If{$\sigma_\mathrm{displaced} > \sigma_\mathrm{tubes}$}{ 
                $\sigma_\mathrm{tubes} \leftarrow \sigma_\mathrm{tubes} +1$  \\
                $\sigma_\mathrm{displaced} \leftarrow 1$  \\
            }
            \lIf{$\sigma_\mathrm{tubes} > \sigma_\mathrm{tubes}^\star$}{
                \textbf{return} $\theta$
            }
            $P.$\setDifficulty{$\sigma_\mathrm{tubes}$, $\sigma_\mathrm{displaced}$}\\ 
        } 
    }
\algorithmfootnote{
    \begin{description} 
         \item \texttt{setDifficulty()} Update the problem difficulty based on ($\sigma_\mathrm{tubes}$, $\sigma_\mathrm{displaced}$). 
         \item \texttt{synchronize()} Synchronize the parameter $\theta$ with the primary neural network every $I_\mathrm{eval}$ seconds.
         \item \texttt{evalPerformance()} Evaluate an agent's performance on a given problem. It returns the number of trials successfully solved by the agent. A trial is deemed successfully solved only if the length of the inferred solution sequence is less than or equal to $H_\mathrm{var}$.
    \end{description}
}
\caption{Curriculum Learning}
\label{alg:cl}
\end{algorithm} 

\subsubsection{Tabu Search}
Within each episode of the actor's exploration, we maintain a record of explored states in order to avoid actions that lead to previously visited states. Particularly, we include the following update to the acceptable actions $\mathcal{A}_\mathrm{acc}(\mathbf{s}_i)$ (recall equation \eqref{eq_af}) for each state:
\begin{equation}
\mathcal{A}_\mathrm{acc}(\mathbf{s}_i)^\dagger= \mathcal{A}_\mathrm{acc}(\mathbf{s}_i) \setminus\{\mathbf{a}|\mathbf{s}_i\overset{\mathbf{a}}{\rightarrow} \mathbf{s}_j\in\mathcal{S}_\mathrm{explored}\}\text{.}
\end{equation}
Here, $\mathcal{S}_\mathrm{explored}$ represents the set of all previously visited states in the current episode. The RL agent will select actions from $\mathcal{A}_\mathrm{acc}(\mathbf{s)}^\dagger$ instead of the original $\mathcal{A}_\mathrm{acc}(\mathbf{s})$ to make use of the tabu and avoid repetition.

Note that there might be special exceptions where all actions lead back to previously visited states. In that case, we temporarily relax the constraint (let $\mathcal{A}_\mathrm{acc}(\mathbf{s)}^\dagger= \mathcal{A}_\mathrm{acc}(\mathbf{s})$) to ensure continued exploration.

\subsection{Inference Method}
\label{sec:Inference Policy}
A greedy selection policy $\pi_{\theta^{\star}}$ is obtained after training the RL model. This policy will then be employed to determine the optimal action sequence on the task level. This is mathematically represented as:
\begin{equation}
\pi_{\theta^{\star}}(\mathbf{s})= \underset{\mathbf{a}\in \mathcal{A}_\mathrm{acc}(\mathbf{s})^{\ddagger\ddagger}}{\mathrm{argmax}}Q(\mathbf{s},\mathbf{a}, \theta^{\star}).
\label{eq_task_inference}
\end{equation}
This equation is iteratively applied to find an action sequence from the current state to the goal.\footnote{ Similar to the sequences explored during training, the sequence generated by the policy $\pi_{\theta^\star}$ may still be incomplete or contain redundancies. For the incomplete sequence, we abandon it and judge task planning as a failure. For potential redundancies, we use an A${}^\star$ trimmer to process the raw sequence before sending it to the motion level for pick-and-place generation.}

The $\mathcal{A}_\mathrm{acc}(\mathbf{s})^{\ddagger\ddagger}$ in equation \eqref{eq_task_inference} is defined as:
\begin{equation}
    \mathcal{A}_\mathrm{acc}(\mathbf{s})^{\ddagger\ddagger} = \mathcal{A}_\mathrm{acc}(\mathbf{s})^\dagger\cap \mathcal{A}_\mathrm{acc}(\mathbf{s})^\ddagger,
\label{eq_close_loop}
\end{equation}
where $\mathcal{A}_\mathrm{acc}(\mathbf{s})^\ddagger$ indicates updates from motion planning failures and is included to take into account the feedback from the motion level.

In the following part, we will show how $\mathcal{A}_\mathrm{acc}(\mathbf{s})^\ddagger$ is updated throughout the planning procedure. Recall the six equations in \eqref{eq_slot_cond}. They consider collisions at the task level and ensure robotic fingers can be posed around a slot without collision. However, they do not take into account motion-level constraints. The task-and-motion planning loop in the framework may become trapped if we simply examine equation \eqref{eq_cond_a_all}. For instance, the task-level agent suggests an action meeting a certain condition, i.e. $C_1(x,y)$. The motion-level planner finds that the robot could not reach any grasp poses associated with $C_1(x,y)$ due to robotic kinematic constraints, obstacles in the workspace, etc., and requests a replanning in the task level. The task-level agent runs again and continues to suggest an action meeting $C_1(x,y)$ and the framework is trapped.

To avoid this trap, we maintain a condition set $\boldsymbol{\boldsymbol{\zeta}}_{xy}$ for each of the rack slots across the different planning levels. In the beginning, all conditions are valid, and the values of $\boldsymbol{\zeta}_{xy}$ are set to $\{1,2,...,6\}$. As the planning progresses, some conditions are invalidated considering the motion-level failures, and their corresponding numbers in the set will be deleted. The task-level agent judges acceptable actions by only checking constraints formed by the remaining conditions denoted by the $\boldsymbol{\zeta}_{xy}$ sets in case of replanning requests. For the previous example, $\boldsymbol{\zeta}_{xy}$ becomes $\{2,3,...,6\}$ after the motion-level planner finds candidate grasp poses belonging to $C_1(x,y)$ cannot be reached. The task-level agent will infer new action sequences by considering $\bigvee_{w=2\sim6} C_{w}(x,y))$ during replanning and thus avoid the trap.

After including $\boldsymbol{\zeta}_{xy}$, the equation for judging an acceptable action becomes
\begin{equation}
\label{eq_ac_pr}
\mathrm{acc}(\mathbf{a})^\prime: (\bigvee_{w\in\boldsymbol{\zeta}_{jk}} C_{w}(j,k)) \land (\bigvee_{w\in\boldsymbol{\zeta}_{pq}} C_{w}(p,q)) \land C_7.
\end{equation}
For a state $\mathbf{s}\in \mathcal{S}$, its set of acceptable actions is defined as
\begin{equation}
\mathcal{A}_\mathrm{acc}(\mathbf{s})^\ddagger = \{\mathbf{a}\ |\ \mathrm{acc}(\mathbf{a})^\prime>0\}.
\label{eq_af_dd}
\end{equation}
This action set is combined with the original $\mathcal{A}_\mathrm{acc}(\mathbf{s})^\dagger$ following equation \eqref{eq_close_loop} to avoid inferring action sequences that are known to be unsolvable in the motion level. Maintaining the condition set $\boldsymbol{\zeta}_{xy}$ for each rack slot allows our framework to effectively perform replanning and quickly find solutions in the presence of failures.

\section{Planning at the Motion Level}
\label{sec:integrating_task_level_rl_and_mp} 
We plan the low-level robot motion based on pick-and-place primitives. Each pick-and-place primitive involves four steps:
1) Approaching the robot hand above the test tube;
2) Grasping and lifting the test tube from the rack;
3) Moving the test tube to its target slot;
4) Inserting the test tube into the target slot and releasing it.
To generate the motion of a pick-and-place primitive, the planner first finds collision-free and IK-feasible grasp poses (reasoning the shared grasp poses) and then plans the detailed joint motion. The details of reasoning and planning will be presented below. 

Note that since a swap action is represented as a one-hot vector, the planner needs to examine the two slots in the action and determine the pick-up slot and place-down slot before working on the primitive. The details of the examination are omitted in the paper due to page limits.

\subsection{Reasoning Shared Grasp Poses}
A test tube must be picked up and placed down using the same grasp pose, as the robot hand and a firmly grasped tube are expected to remain relatively stationary across a single pick-and-place primitive. We call this same grasp pose the ``shared grasp pose'' between the test tube at the pick-up slot and the place-down slot. The shared grasp poses must be identified (reasoned) beforehand so that the motion planner can have multiple candidate grasp poses for generating the successful picking and placing of a test tube. 

\begin{figure}[!htbp]
  \begin{center}
  \includegraphics[width=\linewidth]{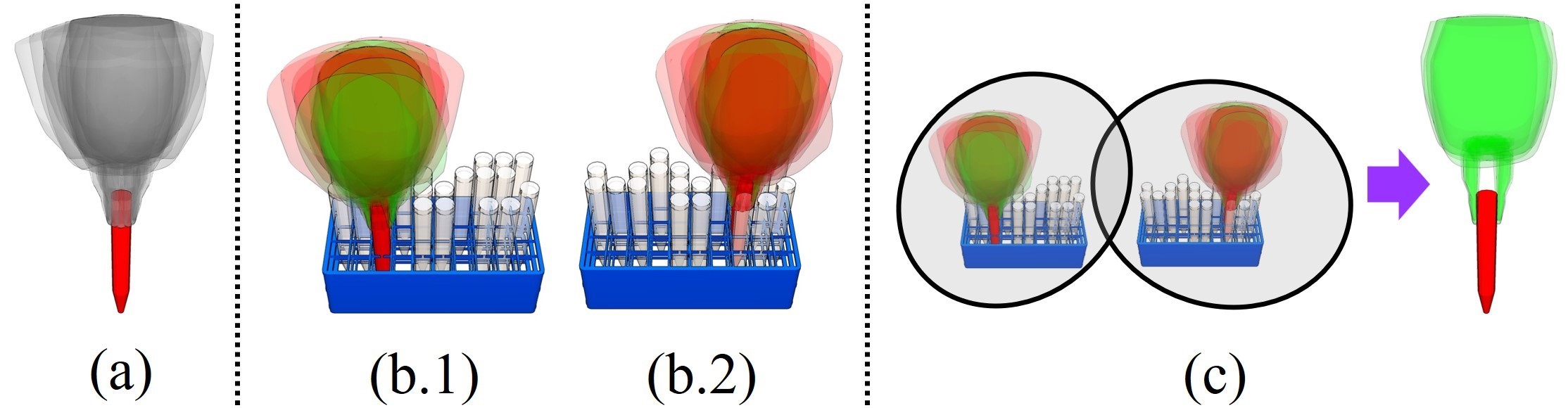}
  \caption{Grasp reasoning. (a) Annotated grasps. Note that grasp poses are down-sampled to be illustrated clearly. (b.1,2) The feasible grasp poses at the pick-up test tube pose and place-down test tube pose. The target test tube is rendered red. The white test tubes are obstacles. The grasp poses are colored to let readers identify that the reasoning results will be from them. The green ones are ``shared grasp poses'' and the red ones are not shared. (c) Reasoning process and results.}
  \label{fig_grasp_reasoning}
  \end{center}
\end{figure}

Fig. \ref{fig_grasp_reasoning} shows an example of reasoning the shared grasp poses. As preparatory work, we manually annotate grasp poses for test tubes. The annotation is performed once for each tube type and can be reused repeatedly. Fig. \ref{fig_grasp_reasoning}(a) exemplifies an annotation result. Reasoning is done based on the annotated grasps. First, we transform the annotated grasp poses to test tubes at the pick-up and place-down slots. Based on the transformed poses, we perform a two-step exclusion process. The first step is conditional exclusion, where we examine the condition sets maintained for the slots and eliminate the poses associated with invalid conditions. The second step is reachability exclusion, where we address robotic inverse kinematics, check collisions, and exclude poses that are unreachable or lead to collisions. The red hands in Fig. \ref{fig_grasp_reasoning}(b.1) and (b.2) show the excluded grasp poses. The remaining grasp poses after the two-step exclusion process are called the ``candidate grasp poses''. The green hands in Fig. \ref{fig_grasp_reasoning}(b.1) and (b.2) illustrate the candidate grasp poses. The shared grasp poses are identified by comparing the candidate grasp poses at the pick-up and place-down slots. The green illustration in Fig \ref{fig_grasp_reasoning}(c) shows the finally identified shared grasp poses in this example.

There are a few points to note for developing a successful reasoning algorithm. (i) Test tubes have rotational symmetry along their longitudinal axis. The grasp poses with the same grasping point but different rotations around this axis should be considered identical when carrying out the reasoning. (ii) The grasping points are recommended to be selected near the test tubes' cap, with the hand's approaching directions having small angles with the tube's longitudinal axis. Such selection helps reduce computational costs as it lowers collision risks with adjacent tubes and prevents the unintentional grasping of multiple tubes. (iii) As illustrated in Fig \ref{fig:conditions}, each condition is associated with a subset of the annotated grasp poses. The conditional exclusion examines and eliminates the transformed grasp poses that are not associated with the conditions in the maintained condition set of a tube's hosting slot. It is fast and helps decrease the number of candidates to be examined by the time-consuming reachability exclusion, thus improving efficiency.

\subsection{Pick-and-Place Motion Planning}
After obtaining the shared grasp poses, we generate the linear lifting and inserting motion using an iterative IK solver, and employ the RRT-connect \cite{kuffner2000rrt} method to plan the motion between the last robot configuration of lifting and the first robot configuration of insertion. If a failure occurs in either lifting, insertion, or the RRT-connect planning, an alternative shared grasp pose is selected for a new attempt. If all shared grasp poses are infeasible, the framework invalidates the maintained condition sets based on the failures and returns to the task level for re-inference and replanning. The invalidation details will be discussed in the next subsection.

\begin{figure}[!htbp]
  \begin{center}
  \includegraphics[width=\linewidth]{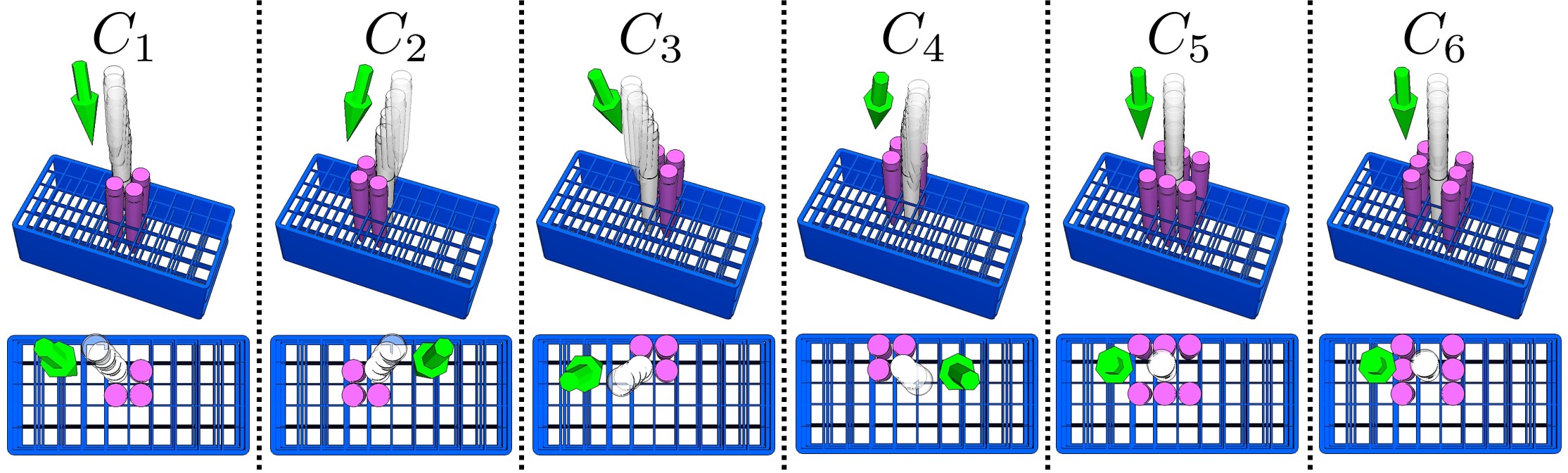}
  \caption{Insertion directions designed for each condition.}
  \label{fig_direction}
  \end{center}
\end{figure}

The insertion directions are crucial to successfully generating the linear motion in the robot workspace. To minimize collision risks, a unique direction is designed for each condition, as illustrated in Fig. \ref{fig_direction}. An average direction is computed based on these directions when multiple conditions are met. For slots satisfying all conditions, a vertical insertion is designated.

\section{Close the Loop by Invalidating the Maintained Condition Sets}
\label{sec:motion_level_replanning}

\subsection{Planning Failures}

If the planning at the motion level fails, the framework will invalidate the maintained condition sets based on the failures and return to the task level for a new inference. In this way, the framework closes the loop while preventing the task-level planner from repeatedly attempting unacceptable actions. There are two particular cases of such invalidation.

\subsubsection{Unreachable Grasp Poses}

As mentioned in Section VI.B, the conditional exclusion filters transformed grasp poses of a tube based on the maintained condition set of its hosting slot. Then, the reachability exclusion follows the conditional exclusion to further examine the IK and collisions of the remaining grasp poses. If none of the remaining grasp poses is reachable, we invalidate the conditions associated with them so that the planner does not judge action accessibility based on them in the next loop. More formally, suppose the remaining grasp poses at $e_{jk}$ after conditional exclusion are associated with a condition set $\boldsymbol{\xi}_{jk}$. When none of these grasp poses are reachable, we will update the maintained condition set of $e_{jk}$ following $\boldsymbol{\zeta}_{jk}=\boldsymbol{\zeta}_{jk}-\boldsymbol{\xi}_{jk}$. The task-level planner will thus select different actions in the next loop according to equations \eqref{eq_close_loop}, \eqref{eq_ac_pr}, and \eqref{eq_af_dd}. 
 
\subsubsection{Environmental Obstruction} 

When environmental obstacles block the robot's motion, the robot cannot move the tube even if solvable kinematic configurations exist. Therefore, we ban all future actions on the tube to avoid useless re-exploration. This can be implemented by invalidating the whole condition set of the affected tube and slot. Assume that a test tube is initially located in a slot $e_{jk}$ and is expected to be moved to a slot $e_{pq}$. If lifting motion is obstructed, we invalidate all conditions in $\boldsymbol{\zeta}_{jk}$ by setting it to $\emptyset$. Then, all task-level actions involving slot $e_{jk}$ become unacceptable and will not be selected in the next loop. Similarly, if insertion motion is blocked, $\boldsymbol{\zeta}_{pq}$ is set to $\emptyset$ and all task-level actions involving slot $e_{pq}$ will be unacceptable in the future. However, if the RRT-connect planning fails, we consider the robot to be restricted and require human intervention to tidy up the environment.

Note that a tube becomes a dead one when the maintained condition set of its hosting slot becomes empty. In that case, the planner will never be able to find a sequence to the goal pattern and will report a fatal failure. In case of fatal failures, human intervention will be required to complete rearrangement tasks. The planner will also report fatal failures if the task planning exceeds the maximum iteration limit without finding a solution.


\subsection{Sensory Failures}
The framework is ready to incorporate various sensory feedback like gripper jaw width, vision, and waist torque/force (motor current) for robust execution. The errors or failures detected by the sensors can be solved in the same way as planning failures within the closed loop.

\subsubsection{Gripper Jaw Width} When a robot gripper grabs a tube with force control, we can monitor the stopping jaw width to ensure the tube is successfully grabbed. If the gripper successfully grabbed the test tube, its jaw width must match the test tube diameter. Otherwise, the grasp should be judged as a failure. Particularly in the case where the gripper fails to grab a tube, in-hand or externally mounted vision may be triggered to re-perceive the workspace and re-analyze the poses of tubes and racks. After that, the motion-level planner will attempt alternative grasp poses and regenerate motion accordingly \cite{chen2023auto}\cite{chen2023rack}. If no alternative grasp poses are available, the system backtracks to the task level for new action sequences.

Another usage of gripper jaw width is to ensure successful tube transfer. When a robot is moving while grasping a tube, we may monitor the gripper jaw width to detect the risk of losing the tubes. If a large disparity between the gripper jaw width and the test tube diameter is found, the framework may be configured to report a slipped or missing tube and request human intervention to resolve the issue before continuing.

\subsubsection{Vision} The robotic pick-and-place motion may cause slight vibrations in the rack and test tubes, leading to accumulated errors and, finally, significant changes in the rack and tube poses. Frequently employing in-hand or external vision to perceive and update the environment during execution helps ensure robustness in the presence of these changes.\footnote{This can be ignored if the environment is strictly structured.} Note that the perceived rack arrangements may differ from planned states in case of failures. The planner may backtrack to the task level to infer new action sequences when such a difference is detected.

The proposed framework accepts two types of vision installation: 1) Installing a camera externally, i.e. above the workspace. 2) Installing a camera on the robot hand, i.e. hand-eye system. For an external camera, a robot's body may occlude observations of the environment. Thus, we only trigger environmental perception after one or multiple pick-and-place actions are finished and a robot moves to a predefined location that does not occlude the camera. Compared with external cameras, a hand-mounted camera is more efficient as it moves with a robot. A hand-mounted camera is seldom occluded, and the hosting robot may continuously monitor the poses of tubes held in the hand with it during execution. The robot may also observe and update rack poses before pick-up or place-down motion, and adjust lifting and insertion directions accordingly.

\subsubsection{Torque/Force (Current)} Robots equipped with T/F or current sensors can monitor for potential collisions and execution failures. When lifting up a test tube, the hand-held tube may collide with surrounding tubes. When inserting a test tube into a rack slot, the hand-held tube may also hit the rack frame or other tubes. By tracking force and torque changes using T/F or current sensors, we can easily detect such collisions and failures. For example, sudden large spikes in force or current signals may be recognized as a failed motion. 

If collisions or failures are detected during lifting, vision feedback may be used for re-perception and replanning. If failure is detected during insertion, further processing is required. For single-arm robots, human intervention might be needed to resolve the insertion failure. For dual-arm robots, the second arm can be employed for automatic correction. A dual-arm robot may use a secondary gripper to grip the tube held by the first hand while keeping the two gripping directions orthogonal. The orthogonal gripping geometrically constrains the tube displacements in non-longitudinal directions. They help eliminate uncertainty in the rotation of the tube, thereby reducing insertion collisions and failures.


\section{Experiments and Analysis}
\label{sec:experiements}
This section presents the experiments conducted to evaluate the efficiency and robustness of our framework. First, we compare the efficiency and success rate of the proposed RL-based task planner with conventional A${}^\star$-based task planner in solving multi-class in-rack test tube rearrangement problems. Second, we evaluate the effectiveness and efficiency of the A${}^\star$ rescuer and A${}^\star$ trimmer in the RL training process. Third, we conduct an ablation study to understand the contributions of each component in our distributed Q-learning structure. Finally, we show real-world experiments and discuss the robustness and adaptability of our framework in a practical scenarios. 

The experiments are performed on a PC equipped with an Intel Core i9-13900KF CPU (3.0 GHz), 64 GB of memory (DDR5, 4800 MHz), and an NVIDIA GeForce RTX 4090 GPU, running Python 3.9. The WRS robotic system\footnote{\url{https://github.com/wanweiwei07/wrs}} is used for motion planning and robot control. 

\begin{figure*}[!htbp]
  \begin{center}
  \includegraphics[width=\linewidth]{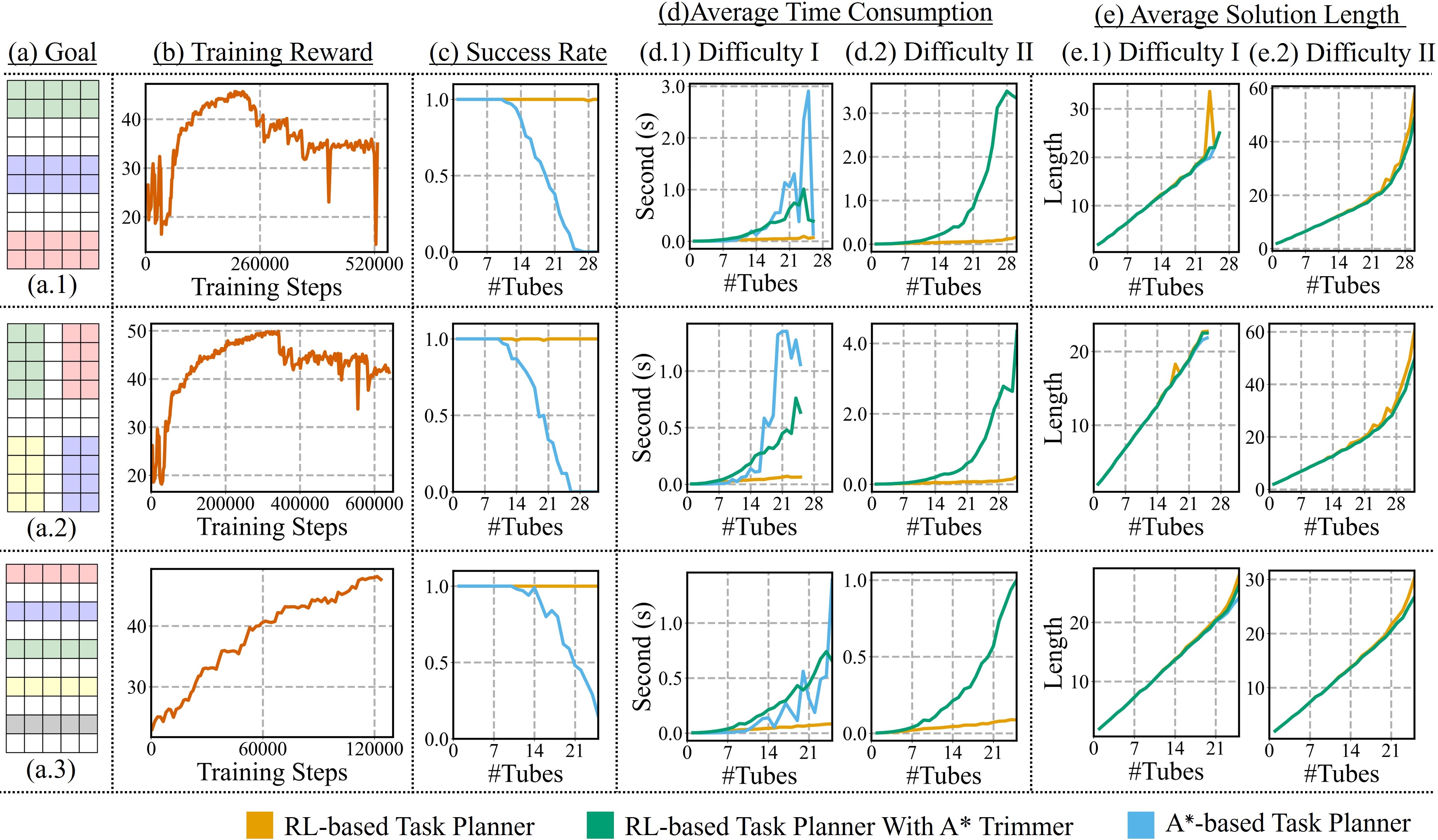}
  \caption{Experimental results for 3 different goal patterns. 'Difficulty I' diagrams are the results of problems that can be successfully resolved by the A${}^\star$ planner. The 'Difficulty II' diagrams are the results of remaining problems.}
  \label{fig:experimental_results}
  \end{center}
\end{figure*}

\subsection{Comparison with A${}^\star$-Based Task Planner}
In this subsection, we present comparative studies among the RL-based task planner and an A${}^\star$-based task planner. We focus on three primary performance metrics: Success rates, time consumption, and solution quality.

As for the problem setup, we assumed 3 to 5 types of tubes and a rack with $5\times10$ slots. We trained three specialist agents for the RL-based planner. Fig. \ref{fig:experimental_results}(a.1)-(a.3) show the goal patterns for these three specialist agents respectively. The average training rewards for the three agents in the curriculum learning evaluator are shown in Fig. \ref{fig:experimental_results}(b). The agents finished curriculum learning at 527000, 639500, and 123500 training steps, respectively. To evaluate the performance, we let the RL-based task planner and A${}^\star$-based task planner solve testing problems with the same random initial states while changing the number of tubes from 1 to the maximum number of test tubes that can match goal patterns. The tube types were randomly assigned while regulating the particular tube number of each type did not exceed the number of slots in the goal. In addition, we let the agents conduct 100 inference or search trials for each number of test tubes to ensure statistical significance.

\subsubsection{Success Rate}
Success rate represents the percentage of successful trials within a maximally allowed number of iteration steps. For the proposed RL-based task planner, the maximum number of iteration steps is set to 300. For the A${}^\star$ heuristic search, the number is set to 1500. Post-processing is ignored in the study here.

Fig. \ref{fig:experimental_results}(c) shows the success rates achieved by both methods as the number of tubes in the initial states increases. The results indicate that the RL-based planner had significantly higher success rates than A${}^\star$ for all specialist agents and tube numbers. The advantage of the RL agent is significantly noticeable when the tube number exceeds 20. While the A${}^\star$ agent success rate dropped to around $60\%$ with 20 tubes and continued decreasing as more tubes were added, the RL agent remained to have around $100\%$ success rate across all trials. For the goal patterns with 3 and 4 tube types (first two rows), when the number of objects exceeded 25, the A${}^\star$ agent failed to find a solution within the maximally allowed iterations. In contrast, the RL agent was not influenced by the increasing tube numbers and remained reliable. The results underscore the RL agent's ability to learn robust policies that can generalize effectively as task complexity increases.

\subsubsection{Time Consumption}
In this part, we compare the time needed by the task planners to find a successful action sequence. To facilitate comparison, we show the results according to the testing problems' levels of difficulty. The 'Difficulty I' problems are those that can be successfully resolved by the A${}^\star$ planner. The 'Difficulty II' problems are the remaining ones. Since 'Difficulty I' problems can be solved by the A${}^\star$ planner, their results have explicit A${}^\star$ curve, as shown by diagrams in the Fig.\ref{fig:experimental_results}(d.1) column. The results of 'Difficulty II' problems do not have an explicit A${}^\star$ curve.

The results in Fig. \ref{fig:experimental_results}(d.1) show that the RL-based planner can solve the rearrangement tasks in under 0.162 seconds and is not much affected by tube numbers. In contrast, the A${}^\star$-based task solver has good performance when the number of tubes is less than 7. However, its time consumption increased exponentially when there were additional tubes. There is some fluctuation in the curve for the A${}^\star$-based planner. This is likely because the randomly generated initial states have varying difficulty. When the tube number is large, the A${}^\star$-based planner can only solve problems with initial states close to the goal patterns. The results in Fig. \ref{fig:experimental_results}(d.2) show that the RL-based planner can scale to more complex problems. The planner's time consumption changed little between the two difficulty levels. However, the A${}^\star$-based planner cannot solve the 'Difficulty II' problems with the given iteration limit.

In addition to the RL-based planner, we also showed the time consumption of ``RL-based planner+A${}^\star$ trimmer'' in Fig. \ref{fig:experimental_results}(d.1) and (d.2). ``A${}^\star$ trimmer'' involves iterative search, and thus, the curve of ``RL-based planner+A${}^\star$ trimmer'' obviously growed with increased tube numbers. Also, as the difficulty level elevated, the A${}^\star$ trimmer required more time to reduce redundancies. The results remind us that we may need to carefully minimize $H_\mathrm{trm}$ to prevent significant trimming costs.

In conclusion, the proposed RL approach enables solving the rearrangement tasks with consistent efficiency for varying tube numbers, which makes it well-suited for applications that need fast response and replanning. Although the A${}^\star$ trimmer slows down efficiency, we still encourage its inclusion to prevent redundancies and thus reduce ineffectiveness in the RL solutions. We will further illustrate this point in the following discussion about average solution length.

\subsubsection{Solution Quality}
Solution quality evaluates the number of actions in solutions required to rearrange tubes into the goal state. The RL-based planner, the RL-based planner with A${}^\star$-trimmer, and the A${}^\star$-based planner are studied and compared in the solution quality evaluation.

Fig. \ref{fig:experimental_results}(e.1) shows the results for ``Difficulty I'' problems. Mostly, the RL-based planner's solutions are comparable to the A${}^\star$-base planner's solutions. However, some fluctuation can be observed when the tube number becomes large. The same tendency is also visible in the ``Difficulty II'' diagrams (Fig. \ref{fig:experimental_results}(e.2)). The fluctuation occurred because, during curriculum learning, we prioritized completion over solution length as the success criteria to avoid getting stuck. Carefully designing the success criteria during training could reduce the length of inferred sequences but would require significant human effort.

The A${}^\star$-trimmer demonstrated to be an effective way to suppress the fluctuation. This can be observed from the results of RL-based planner with A${}^\star$-trimmer in both Fig. \ref{fig:experimental_results}(e.1) and (e.2). The curves became smoother and shorter compared with those of the singular RL-based planner.

\subsection{Effectiveness of the A${}^\star$ Rescuer and Trimmer }
In this subsection, we evaluate the effectiveness of the A${}^\star$ rescuer and trimmer in RL training. We particularly focused on the RL training for the goal pattern shown in Fig.\ref{fig:experimental_results}(a.3), and compared four different combinations: (i) RL with both the rescuer and trimmer, (ii) RL with only the rescuer, (iii) RL with only the trimmer, and (iv) singular RL. These combinations are denoted as ``Baseline + R. + T.'', ``Baseline + R.'', ``Baseline + T.'', and ``Baseline'' respectively for short. By comparison, we aim to ascertain the individual and conjoint impacts of the rescuer and the trimmer on the training process. All four combinations employed identical parameters for distributed Q-learning and D3QN networks. 

\begin{figure}[!htbp]
  \begin{center}
  \includegraphics[width=\linewidth]{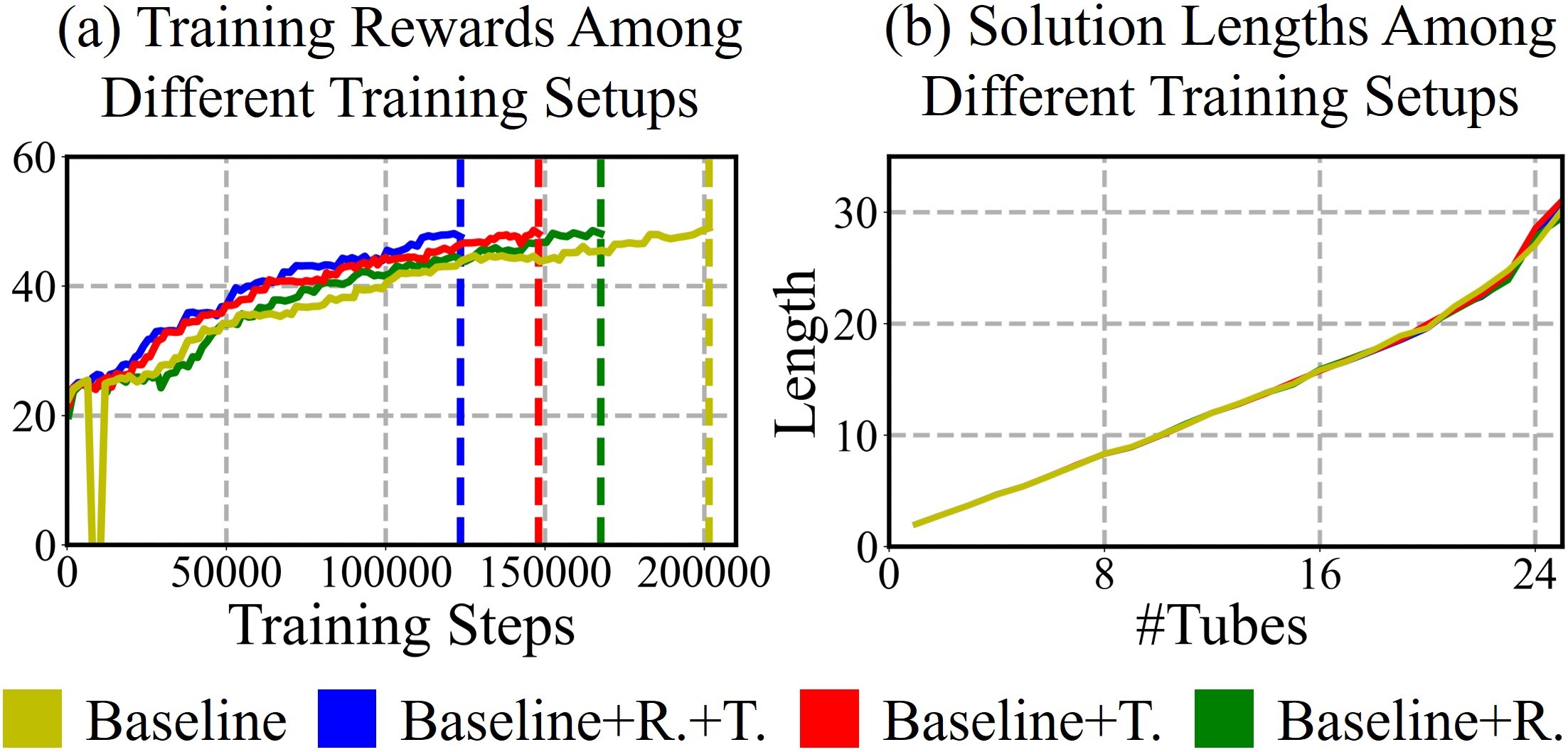}
  \caption{(a) Average rewards obtained in the evaluator of distributed Q-learning. The dashed horizontal line indicates when agents finished curriculum learning across different RL training setups. (b) The average solution lengths for the agents trained by different RL setups. For each number of test tubes, $100$ trials were evaluated with random initial starts. All agents with different setups achieved a $100\%$ success rate in solving these trials.}
  \label{fig:comp_1}
  \end{center}
\end{figure}

Fig. \ref{fig:comp_1}(a) presents the average rewards obtained by the different combinations during training and also the training steps needed to complete the curriculum. The results show that ``Baseline + R. + T.'', ``Baseline + R.'', and ``Baseline + T.'' complete the curriculum learning respectively $38.7\%$, $16.9\%$, and $26.6\%$ faster than the baseline, which verifies the effectiveness of our concept of using heuristics in a post-processing procedure to amplify the data. The combination of both the rescuer and trimmer is especially distinct and has the best convergence speed. The results also show that ``Baseline + R. + T.'' and ``Baseline + T.'' outperformed ``Baseline + R.'' and "Baseline". This implies that agents with the A${}^\star$ trimmer helped focus more on actions that are more likely to contribute to the goal. Fig. \ref{fig:comp_1}(b) compares the solution quality of agents trained with the different combinations using different numbers of tubes. The results show that the various combinations achieved similar action sequence lengths, which implies that the acceleration methods do not cause bias or compromise the agents' scalability. 

Table \ref{tab:data_reduces} presents the detailed results about the completion and reduction achieved by the A${}^\star$ rescuer and trimmer. In ``Baseline + R. + T.'', the rescuer saved $85\%$ of the $392$ incomplete action sequences, and the trimmer reduced $30.7\%$ of total actions. In ``Baseline + T.'', the trimmer reduced $29.3\%$ of total actions.

\begin{table}[!htbp]
\caption{Detail results of the A${}^\star$ rescuer and trimmer.}
\label{tab:data_reduces}
\centering
\begin{threeparttable}
\begin{tabular}{l|rcl|rcl}
\toprule  
Combination & \multicolumn{3}{c|}{\#Saved / \#Incomplete} & \multicolumn{3}{c}{\#Reduced / \#Total}\\
\midrule
Baseline + R. + T. & 334 & / & 392 & 264002 & /  & 859805  \\
Baseline + R. & 3677& / & 4318 &  0 & / & 8000860  \\
Baseline + T. & 0 & / & 440 &  314287 & / & 1073018  \\
\bottomrule
\end{tabular}
\begin{tablenotes}
  \item[Note] \#Saved: Actions completed by the rescuer. \#Incomplete: Number of incomplete action sequences. \#Reduced: Actions reduced by the trimmer. \#Total: Total number of actions.
  \end{tablenotes}
\end{threeparttable}
\end{table}

\subsection{Effectiveness of the Distributed Q-Learning Structure}
In this subsection, we study the contribution of each component in the proposed distributed Q-learning structure, with special attention paid to the importance of curriculum learning and post-processing. Like the previous subsection, the RL agents in the evaluation are trained for the goal pattern shown Fig. \ref{fig:experimental_results}(a.3).

\subsubsection{Varying Component Combinations}
Our distributed Q-learning structure used post-processing and curriculum learning to improve learning performance. In this part, we evaluate the importance of the components by excluding one or both. The new combinations after exclusion include (i) ``Proposed - P.P.'', where the post-processing component is excluded, (ii) ``Proposed - C.L.'', where the curriculum learning is excluded, and (iii) ``Proposed - (P.P. + C.L.)'', where both post-processing and curriculum learning are excluded.

\begin{figure}[!htbp]
  \begin{center}
  \includegraphics[width=\linewidth]{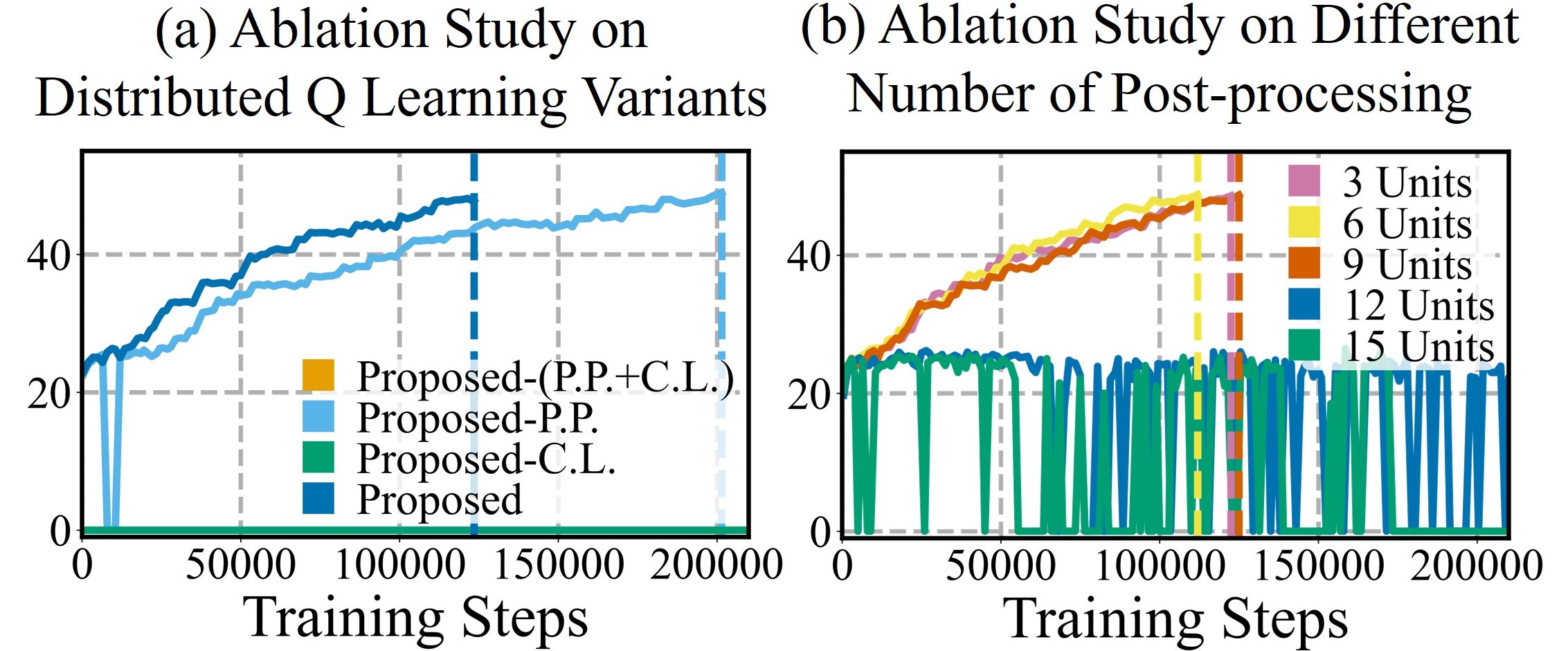}
  \caption{(a) Result of different distributed Q-learning component combinations. (b) Influence of the number of units in the post-processing component.}
  \label{fig:comp}
  \end{center}
\end{figure}

Fig. \ref{fig:comp} presents the average rewards obtained during training for each combination. The results show that: (i) Both ``Proposed - (P.P. + C.L.)'' and ``Proposed - C.L.'' failed to converge, which highlights the essential role of curriculum learning in stabilizing and guiding the learning process. (ii) Post-processing facilitates faster convergence, as evidenced by the curves of ``Proposed'' versus ``Proposed - P.P.''. (iii) The proposed structure is most effective and underscores the conjoint benefits of the combination.

\subsubsection{Number of Post-Processing Units}
The post-processing component includes a number of units that run in parallel to process the sequences obtained by the actors. In this part, we studied how the number of post-processing units in the distributed structure influences agent performance. 

Fig. \ref{fig:comp}(b) shows the average training rewards when the number of post-processing units differs. The results show that 6 post-processing units prompted the highest convergence speed. The agents with either 3 or 9 post-processing units also exhibited comparable convergence. However, when the number reached or exceeded $12$, the agents failed to converge.

We believe the reason is that a large number of post-processing units resulted in imbalanced reward distribution in the replay buffer. The post-processed sequences had positive rewards. Too many post-processing units led to a substantially large portion of positive rewards through the training process, and the RL agent was not able to receive enough examples of non-rewarding or negatively rewarding actions (sparse negative rewards). The trained agent thus got a skewed understanding of the environment where the likelihood of receiving a reward was overestimated for any given action. 

\subsection{Real-World Evaluation}
 
\begin{figure}[!htbp]
  \begin{center}
  \includegraphics[width=\linewidth]{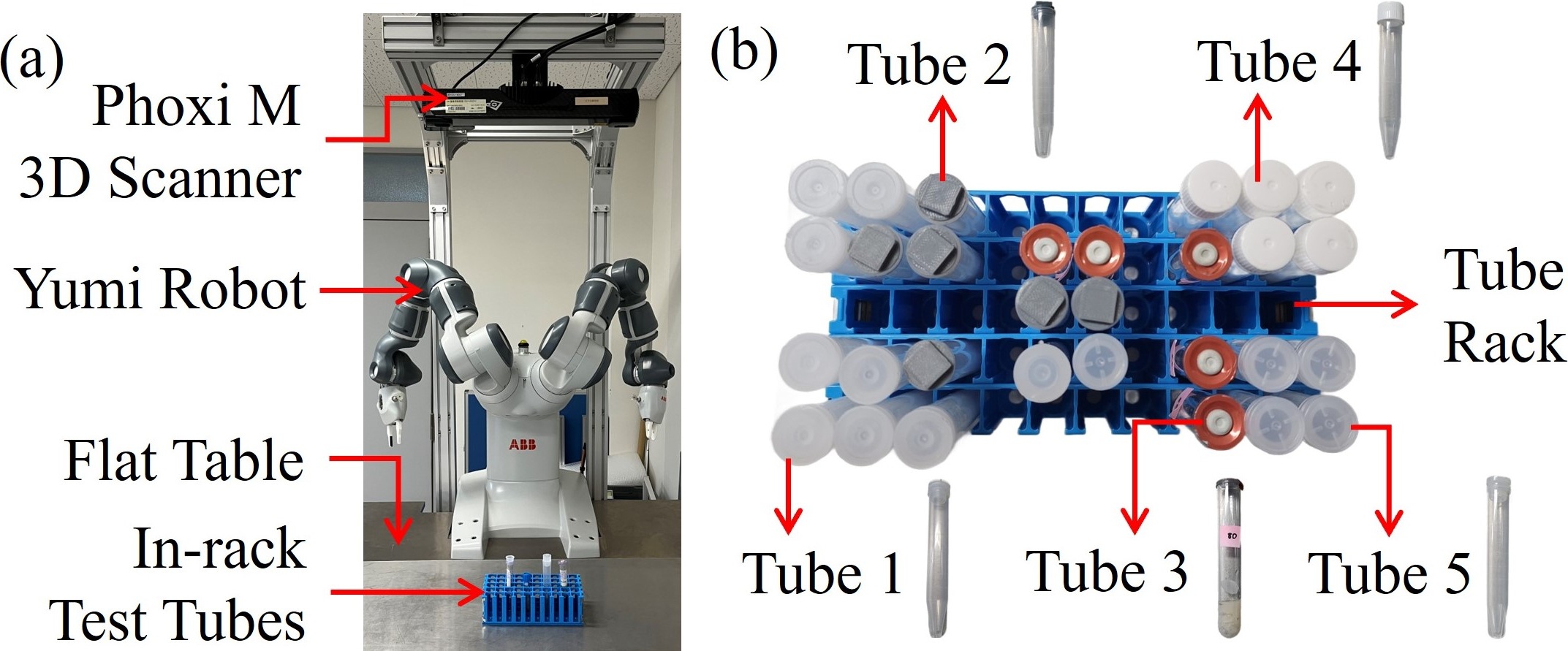}
  \caption{(a) Robot platform used in the test. (b) Tubes and rack in the view of the Phoxi M 3D Scanner.}
  \label{fig:exp_env}
  \end{center}
\end{figure}

Fig. \ref{fig:exp_env} shows the robot platform used for real-world evaluation. It comprises an ABB Yumi dual-arm robot with a two-finger gripper and a Photoneo Phoxi M 3D Scanner for recognizing test tubes and racks on a flat table. A universal tube rack and five types of test tubes are used in the experiment. The visual recognition method presented in \cite{chen2023auto} is leveraged for tube detection. The recognition method can achieve a $98.8\%$ success rate, and thus, the visual uncertainty can be ignored.

\subsubsection{Results and Analysis}
In total, we carried out 20 real-world trials. The first ten used the specialist agent trained on the goal pattern shown in Fig.\ref{fig:experimental_results}(a.1). We randomly assigned 27 (9 for each type) tubes to the initial states of these trials. The later ten trials used the agent trained on the goal pattern in Fig.\ref{fig:experimental_results}(a.3). We randomly assigned 25 (5 for each type) tubes to the initial states of these trials. We assume there are no obstacles on the workbench in front of the robot. Before execution, we dropped the rack into a visible and reachable area of the workbench so that test tubes were accessible. The overhead Phoxi Scanner captured a view of the rack, recognized the initial tube arrangements, and sent them as input to our framework. The robotic execution was performed with two types of sensory feedback, including (i) external vision, and (ii) force/torque from current sensors embedded in the robot's joints. For the vision feedback, when the robot gripper failed to grasp a tube, the robot moved to a location that did not occlude the camera and let the sensor capture the rack again and update the poses of various objects. We adopted this lazy policy instead of frequently re-perceiving the environment after each pick-and-place to avoid wasting time on redundant avoidance actions.\footnote{Note this lazy policy increased failure rates a bit. The tubes might tilt slightly in the rack. Vibrations caused by robotic pick-and-place could increase the tilting uncertainty and occasionally result in weak grasping. Tubes might slip inside the gripper in the presence of such weak grasping.}

\begin{figure*}[!htbp]
  \begin{center}
  \includegraphics[width=\linewidth]{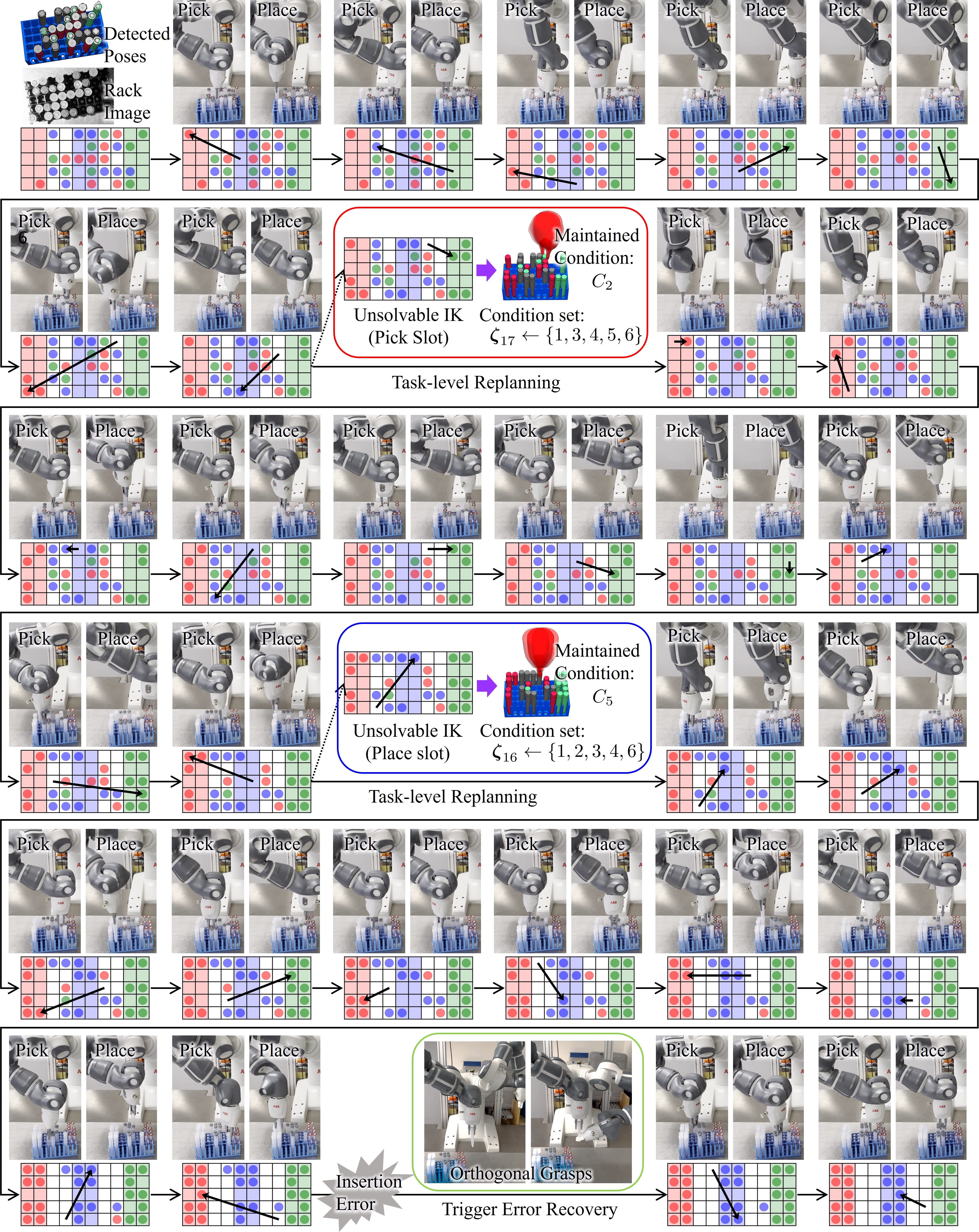}
  \caption{Planning and replanning}
  \label{fig:robot_motions}
  \end{center}
\end{figure*}

An exemplary result is shown in Fig. \ref{fig:robot_motions}. The system perceived the environment and detected the test tubes in the rack. Then the RL-based task planner inferred an action sequence to a goal state. After that, the motion planner converted the sequence to detailed robot motion. The red and blue rounded frames in Fig. \ref{fig:robot_motions} show two motion planning failures. The motion planner failed to find IK-feasible grasp poses, and the associated condition was removed from the slot's maintained condition set. After that the task planner replanned. The green rounded frame in Fig. \ref{fig:robot_motions} shows an insertion failure detected by the force/torque sensor. The robot corrected poses using a secondary gripper to recover from the failure.

Table \ref{tab:time_consumption} provides a detailed breakdown of the experimental results. From the column perspective, the table has three sections. The left section shows the success or failures of each trial. The middle section shows the number of executions for the vision, task planning, and motion planning modules. The right section shows the number of feedback and replanning occurrences within the task and motion planning. From the row perspective, the table is divided into two sections. The upper section presents results for the goal pattern shown in Fig.\ref{fig:experimental_results}(a.1), while the lower section presents the results for the goal pattern in Fig.\ref{fig:experimental_results}(a.3). 

\begin{table}[!htbp]
\caption{Details of the Real-World Results}
\label{tab:time_consumption}
\centering
\begin{threeparttable}
\begin{tabular}{lc|ccc|cccc}
\toprule  
ID & Succ. & \#V. & \#T.P. & \#M.P. & \#G.F. & \#I.F. & \#F.F. & \#R. \\
\midrule
1-1 & True   &2 &1 &44 &1 &0 &0/0/0 &1 \\ 
1-2 & True   &1 &1 &25 &0 &0 &0/0/0 &0 \\ 
1-3 & True   &2 &1 &26 &1 &0 &0/0/0 &1 \\ 
1-4 & True   &1 &4 &31 &0 &1 &0/0/0 &3 \\ 
1-5 & True   &2 &1 &23 &1 &0 &0/0/0 &1 \\ 
1-6 & True   &3 &1 &41 &2 &0 &0/0/0 &2 \\ 
1-7 & True   &2 &1 &35 &1 &0 &0/0/0 &1 \\ 
1-8 & True  & 1 &1 &31 &0 &0 &0/0/0 &0 \\  
1-9 & False  & 1 &1 &23 &0 &0 &0/1/0 &0 \\ 
1-10 & False &1 &3 &6 &0 &0 &1/0/0 &0 \\ 
\midrule
2-1 & True &1 &2 &29 &0 &1 &0/0/0 & 1 \\
2-2 & True &1 &1 &20 &0 &0 &0/0/0 &0 \\
2-3 & True &2 &1 &21 &1 &0 &0/0/0 &0 \\
2-4 & True &2 &1 &29 &1 &0 &0/0/0 &0 \\
2-5 & True &2 &2 &23 &1 &0 &0/0/0 &1 \\
2-6 & True &1 &3 &39 &0 &0 &0/0/0 &2 \\
2-7 & True &1 &1 &12 &0 &0 &0/0/0 &0 \\
2-8 & False &1 &1 &5 &0 &0 &1/0/0 &0 \\
2-9 & False &1 &1 &15 &0 &1 &0/0/1 &0 \\
2-10 & False &2 &3 &27 &1 &3 &1/0/0 &2 \\
\bottomrule
\end{tabular}
\begin{tablenotes}
  \item[Note 1] Succ.: Indicate if a trial is successful. \#V.: Times of visual detection triggered during execution. \#T.P.: Times of task-level inference. \#M.P.: Times of motion planning. \#F.G.: Times of failures in grabbing a tube. \#C.I.: Times of collisions detected during insertion. \#R.: Times of backtracking to the task level for replanning. \#F.F.: Fatal failures.
  \item[Note 2] The upper parts of the table are experimental results for the goal pattern Fig. \ref{fig:experimental_results}(a.1). The lower parts of the table are experimental results for the goal pattern Fig. \ref{fig:experimental_results}(a.3).
  \end{tablenotes}
\end{threeparttable}
\end{table}  
The success rates for completing the trials of these two goal patterns are $80\%$ and $70\%$, respectively. The first goal pattern required more average times of motion planning ($28.5$ vs. $22$, as can be seen from the sum of the \#M.P. column) and was more difficult than the second goal pattern. The robot failed to grab a tube at 1-1, 1-3, 1-5, 1-6, 1-7, 2-3, 2-4, 2-5, 2-10, as denoted by the \#G.F. column. The system triggered re-perceiving for the environment to update the test tube and rack poses in the presence of these failures, and the motion planner then re-generated a pick-and-place motion using other shared grasp poses. Tubes collided with racks when being inserted in trials 1-4, 2-1, 2-9, and 2-10, as denoted by the \#I.F. column. When such collisions were detected, the robot performed the dual-arm correction shown in the green rounded frame of Fig. \ref{fig:experimental_results}. The replanning was triggered in trials 1-1, 1-3, 1-4, 1-5, 1-6, 1-7, 2-1, 2-5, 2-6, 2-10, as denoted by the \#R. column. In these cases, all shared grasp poses were IK-infeasible and the planner backtracked to the task level to re-generate the action sequence. The \#F.F. column shows the fatal failures that led to failed trials. We encountered three kinds of fatal failures in the trials. Their details will be discussed in the next part.

\subsubsection{Discussions on Fatal Failures}
\begin{figure}[!htbp]
  \begin{center}
  \includegraphics[width=\linewidth]{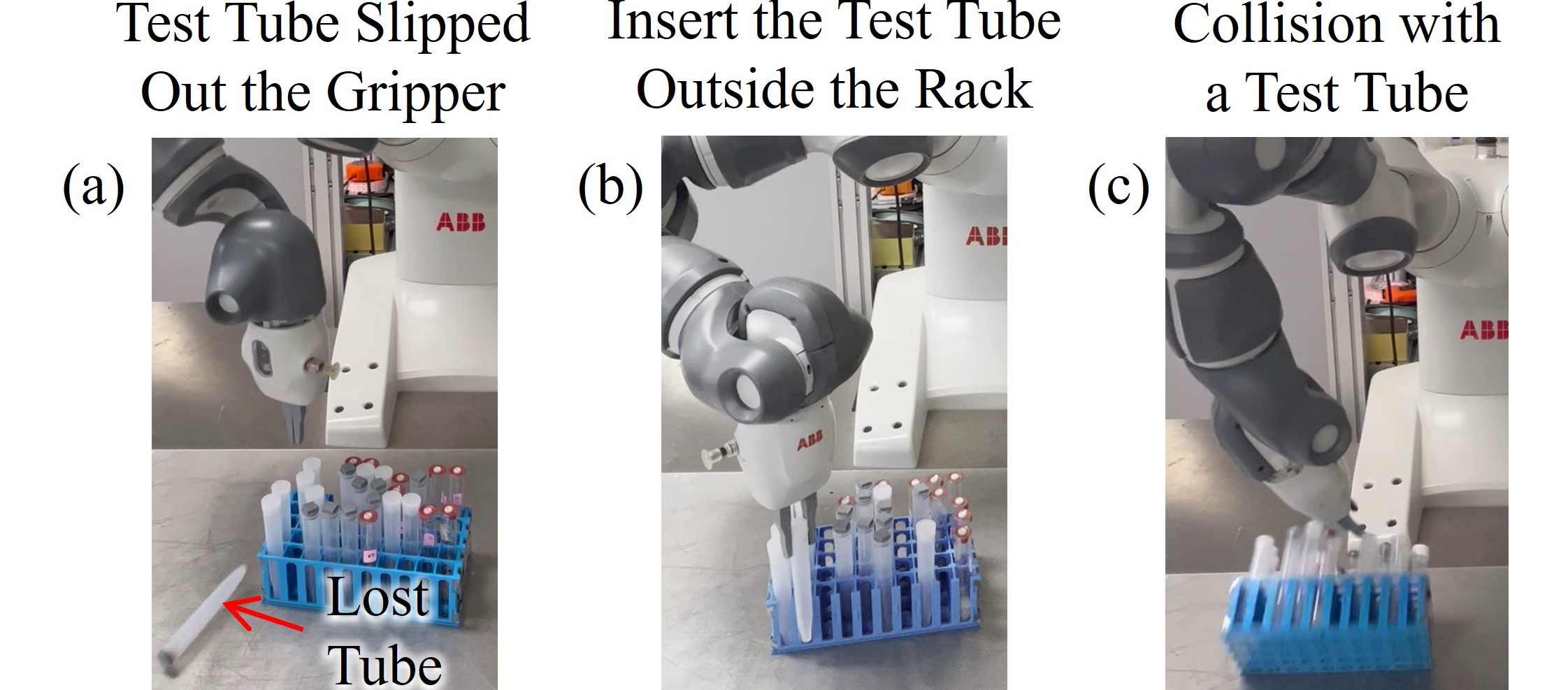}
  \caption{Fatal failures encountered in real-world experiment.}
  \label{fig:err}
  \end{center}
\end{figure}

When fatal Failures happen, the robot can no longer recover and has to request human intervention. We mainly observed three kinds of fatal failures during execution. (i) The test tube slipped out of the gripper. (ii) The test tube slipped inside the gripper and was inserted outside the rack. (iii) Toppling, knocking, or emergent collisions. Fig. \ref{fig:err} shows the pictures taken when these fatal failures were seen. The three values of Table \ref{tab:time_consumption} \#F.F. column show the times of these failures in the trials respectively.

The first fatal error occurred when the robot was moving test tubes. The gripper did not always hold the test tubes firmly due to uncertainties during grasping, and a tube might occasionally slip out of the gripper. In that case, the test tube would be lost and never be correctly positioned. The second fatal error happened when a test tube slipped but was still inside the gripper, and the robot was trying to insert it into a boundary slot. In that case, the tube could be inserted down without collision. The robot would misjudge that the tube was successfully inserted and thus release it. The tube then got dropped outside the rack and could never be correctly positioned. The third fatal error happened when the robot collided with test tubes or racks during its movement. The collision might cause the rack to topple over, knock the rack outside the reachable area, or trigger an emergency brake in the robot. We suggest increasing the sampling distance in the RRT-connect algorithm to reduce the likelihood of such collisions.

\begin{figure}[ht]
\centering
    \begin{minipage}[c]{0.3\linewidth}
        \includegraphics[width=\linewidth]{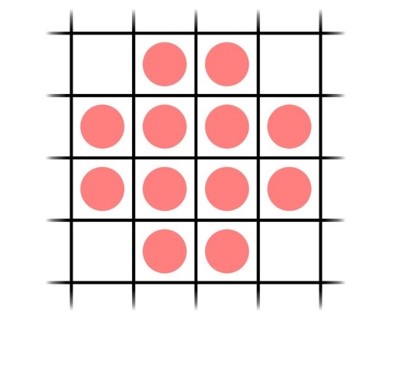}
    \end{minipage}%
    \hspace{0.01\linewidth}
    \begin{minipage}[c]{0.67\linewidth}
        \caption{Example of a ``dead'' arrangement. Test tubes (red circles) inside the pattern obstruct each other, and the task-level planner cannot find a valid grasp pose to pick them.}
        \label{fig:dead_pattern}
    \end{minipage}
\end{figure}

Besides the fatal failures during execution, we also observed task-level failures. As shown in Fig. \ref{fig:dead_pattern}, the RL task planner sometimes encountered ``dead'' arrangements where no valid action meets the constraints formed by the slot's conditions. While our RL agents avoided learning these arrangements, they may initially exist in the starting state and the framework had to report a fatal failure. Grippers with thinner fingers will have relaxed finger constraints and may help solve the `dead'' arrangement failure.

\section{Conclusions}
\label{sec:conclusions}
We proposed the combined task-level RL and motion planning framework to address the multi-class in-rack test tube rearrangement problem. The RL-based task planner was trained efficiently with the help of the distributed structure, curriculum learning, and A${}^\star$ post-processing. It could efficiently find an action sequence to rearrange the test tubes. The RL-based planner was connected to motion planning in a closed loop by maintaining a condition set for each rack slot to track and deal with failures. The closed loop allowed the framework to perform replanning and effectively find solutions in the presence of low-level failures. The framework could also incorporate various sensory feedback for robust execution. Simulations and real-world studies demonstrated the efficiency and reliability of the proposed framework.


\section*{Appendix}
Compared with the original DQN \cite{mnih2013playing}, D3QN incorporates a double Q-learning algorithm to mitigate the overestimation bias inherent to traditional Q-learning methods \cite{van2016deep}, and employs a dueling network architecture to separately estimate the state value function and the advantage function. This separation enhances the network's efficiency in scenarios where multiple actions have similar state-action values, allowing for a more precise determination of the most beneficial action of a given state. 
 
Particularly, the parameters of a D3QN are updated by minimizing the following loss over the transition tuples $(\mathbf{s}_t, \mathbf{a}_t, r_t, \mathbf{s}_{t+1})$ sampled from a replay buffer:
\begin{align}
\label{eq:q_network_update}
    \mathcal{L}(\theta) =~&Q(\mathbf{s}_t, \mathbf{a}_t;\theta) -r_t\nonumber\\
    &-\gamma Q(\mathbf{s}_{t+1}, \underset{\mathbf{a}\in \mathcal{A}_\mathrm{acc}(\mathbf{s}_{t+1})}{\mathrm{argmax}}Q(\mathbf{s}_{t+1},\mathbf{a};\theta);\theta'),
\end{align}
where $\theta$ represents the parameters of the primary Q-network, and $\theta'$ represents the parameters of the target network that is periodically updated. The Q-value $Q(\mathbf{s},\mathbf{a};\theta)$ is computed using a state value function $V(\mathbf{s};\theta)$ and advantage function $A(\mathbf{s},\mathbf{a};\theta)$ according to:
\begin{align}
    Q(\mathbf{s}, \mathbf{a};\theta)=~&V(\mathbf{s};\theta) + A(\mathbf{s},\mathbf{a};\theta) \nonumber\\
    & - \frac{1}{|\mathcal{A}_\mathrm{acc}(\mathbf{s})|}\sum_{\mathbf{a}^\prime\in\mathcal{A}_\mathrm{acc}(\mathbf{s})}A(\mathbf{s},\mathbf{a}^\prime;\theta)\text{.}
\end{align}  
By distinguishing the Q-value between the expected value of a state and the specific advantages of individual actions within that state, the dueling network provides a clearer understanding of the optimal actions to take in various situations. In addition, equation $\eqref{eq:q_network_update}$ involves a subtraction operation that is essential to maintaining a stable learning process. The subtraction helps center the advantage values and thus prevents the advantage values from becoming too large or too small relative to the value function.

\begin{figure}[!htbp]
  \begin{center}
  \includegraphics[width=\linewidth]{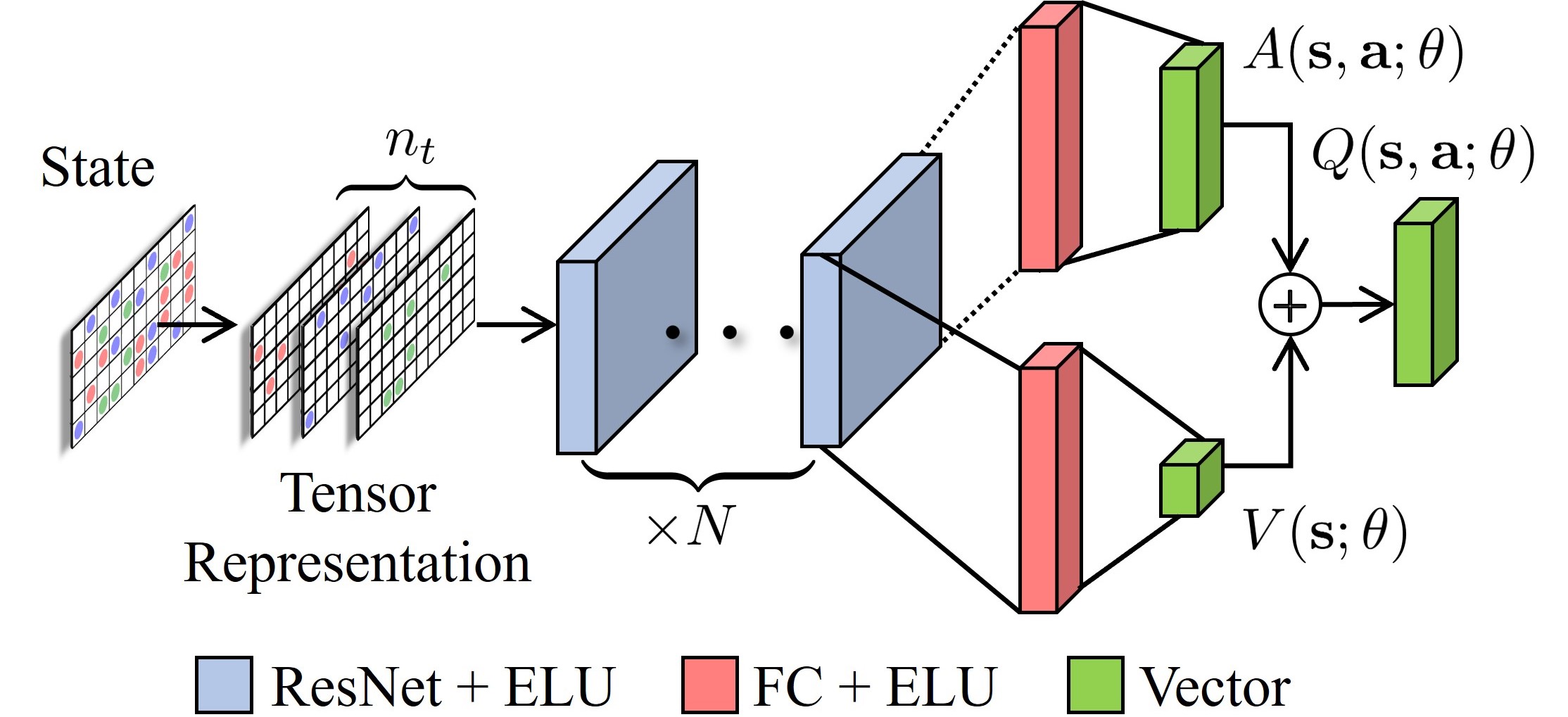}
  \caption{Adopted D3QN architecture. The backbone uses $N$ residual networks (ResNets) \cite{he2016deep} with Exponential Linear Unit (ELU) \cite{clevert2015fast} activation functions. Both the advantage and value heads are fully connected with the ELU activation functions.}
  \label{fig:d3qn}
  \end{center}
\end{figure}

Fig. \ref{fig:d3qn} shows the structure of the D3QN network used in our framework. The hyperparameters for the D3QN architectures are presented in Table \ref{tab:d3qn_parameters}. Note that when inputting a state into the network, we separate each distinct test tube type into individual channels and transform a state's matrix representation into a ($n_t\times n_r \times n_c$)-dimensional tensor representation. The tensor representation helps enable more effective learning. A similar representation can be found in AlphaZero \cite{silver2018general} to encode black and white pieces in chess.

\begin{table}[!htbp]
\caption{D3QN parameter table}
\label{tab:d3qn_parameters}
\centering
\begin{threeparttable}
\begin{tabular}{c|cccc}
\toprule  
Layer & Name & Parameters & Output Size \\
\midrule
 $0$ & Input & - & $(3, 5, 10)$   \\
 \midrule
 \multirow{2}{*}{1} & Convolution & $\mathrm{Conv2d}(48, 3, 1, 1)$ & $(48, 5, 10)$\\
                    & ELU & $\mathrm{ELU}(\alpha =0.1)$ & $(48, 5, 10)$ \\ 
\midrule
 $2\sim8$ & ResNet & 
    $\begin{bmatrix}
        \mathrm{Conv2d}(48, 3, 1, 1) \\
        \mathrm{ELU}(\alpha=1.0)\\
       \mathrm{Conv2d}(48, 3, 1, 1)  \\
    \end{bmatrix}\times 6$
 &  $(48, 5, 10)$  \\
 \midrule
 $9$  & Advantage & $\begin{bmatrix}
        \mathrm{Conv2d}(30, 1, 1, 0) \\
        \mathrm{ELU}(\alpha=1.0)\\
        \mathrm{Flatten}()\\
       \mathrm{Linear}(1225)  \\
    \end{bmatrix}$  &  $(1225)$  \\
\midrule
$10$ & Value & $\begin{bmatrix}
        \mathrm{Conv2d}(10, 1, 1, 0) \\
        \mathrm{ELU}(\alpha=1.0)\\
        \mathrm{Flatten}()\\
       \mathrm{Linear}(1)  \\
    \end{bmatrix}$  &   $(1)$\\
 
\bottomrule
\end{tabular}
\begin{tablenotes}
  \item[Note] The D3QN model uses Pytorch-style modules. The $\mathrm{Con2d}$ function represents a 2D convolutional layer with the following parameters: (i) filters, (ii) kernel size, (iii) stride, and (iv) padding. The parameter for the $\mathrm{Linear()}$ function specifies the output dimension of the linear layer, i.e. the number of neurons.
  \end{tablenotes}
\end{threeparttable}
\end{table}

\bibliographystyle{IEEEtran}
\bibliography{citations.bib}

\end{document}